\documentclass[10pt,twocolumn,letterpaper]{article}

\usepackage[pagenumbers]{cvpr} 

\definecolor{cvprblue}{rgb}{0.21,0.49,0.74}
\usepackage[pagebackref,breaklinks,colorlinks,allcolors=cvprblue]{hyperref}
\usepackage{caption}
\usepackage{float}
\usepackage{multirow}
\usepackage{xcolor} 
\usepackage{booktabs,makecell,multirow,rotating,siunitx,xcolor,colortbl}
\usepackage{amssymb}
\usepackage{tikz}

\usepackage{tabularx}

\def\paperID{12368} 
\def\confName{CVPR}
\def\confYear{2026}

\title{Reasoning Path and Latent State Analysis for Multi-view Visual Spatial Reasoning: A Cognitive Science Perspective}

\author{Qiyao Xue\footnotemark[1] , Weichen Liu\footnotemark[1] , Shiqi Wang\footnotemark[1] , Haoming Wang\footnotemark[1] , Yuyang Wu\footnotemark[2] , Wei Gao\footnotemark[1]\\
	University of Pittsburgh\footnotemark[1] , Carnegie Mellon University\footnotemark[2]\\
	{\tt\small \{qiyao\_xue, weichenliu, shw322, hw.wang\}@pitt.edu, yuyangwu@andrew.cmu.edu, weigao@pitt.edu}
}

\begin{document}
\makeatletter
\g@addto@macro\@maketitle{
	
}
\makeatother
\maketitle

\begin{abstract}
Spatial reasoning is a core aspect of human intelligence that allows perception, inference and planning in 3D environments. However, current vision-language models (VLMs) struggle to maintain geometric coherence and cross-view consistency for spatial reasoning in multi-view settings. We attribute this gap to the lack of fine-grained benchmarks that isolate multi-view reasoning from single-view perception and temporal factors. To address this, we present ReMindView-Bench, a cognitively grounded benchmark for evaluating how VLMs construct, align and maintain spatial mental models across complementary viewpoints. ReMindView-Bench systematically varies viewpoint spatial pattern and query type to probe key factors of spatial cognition. Evaluations of 15 current VLMs reveals consistent failures in cross-view alignment and perspective-taking in multi-view spatial reasoning, motivating deeper analysis on the reasoning process. Explicit phase-wise analysis using LLM-as-a-judge and self-consistency prompting shows that VLMs perform well on in-frame perception but degrade sharply when integrating information across views. Implicit analysis, including linear probing and entropy dynamics, further show progressive loss of task-relevant information and uncertainty separation between correct and incorrect trajectories. These results provide a cognitively grounded diagnosis of VLM spatial reasoning and reveal how multi-view spatial mental models are formed, degraded and destabilized across reasoning phases. The ReMindView-Bench benchmark is available at \url{https://huggingface.co/datasets/Xue0823/ReMindView-Bench}, and the source codes of benchmark construction and VLM reasoning analysis are available at \url{https://github.com/pittisl/ReMindView-Bench}.
\end{abstract}    
\vspace{-0.1in}
\section{Introduction}
Spatial reasoning is essential to infer spatial relations, geometric configurations and dynamics among real-world objects that are not directly observable \cite{freksa2006spatial, tenderra2025human}, thereby supporting higher-level cognitive functions such as navigation and manipulation \cite{ku2020room, liu2023visual}. In particular, spatial reasoning in multi-view visual settings, where reasoning needs information from multiple viewpoints, 
is more difficult, because it requires maintaining geometric coherence and cross-view consistency across different viewpoints \cite{hong20233d, yu2025far,yeh2025seeing} to reconstruct scene layouts and reason about objects' relative positions, orientations and occlusions \cite{das2018embodied,hong20233d,yu2019multi,yang2025thinking}.

\begin{figure*}
	\centering
	\vspace{-0.35in}
	\includegraphics[width=\textwidth]{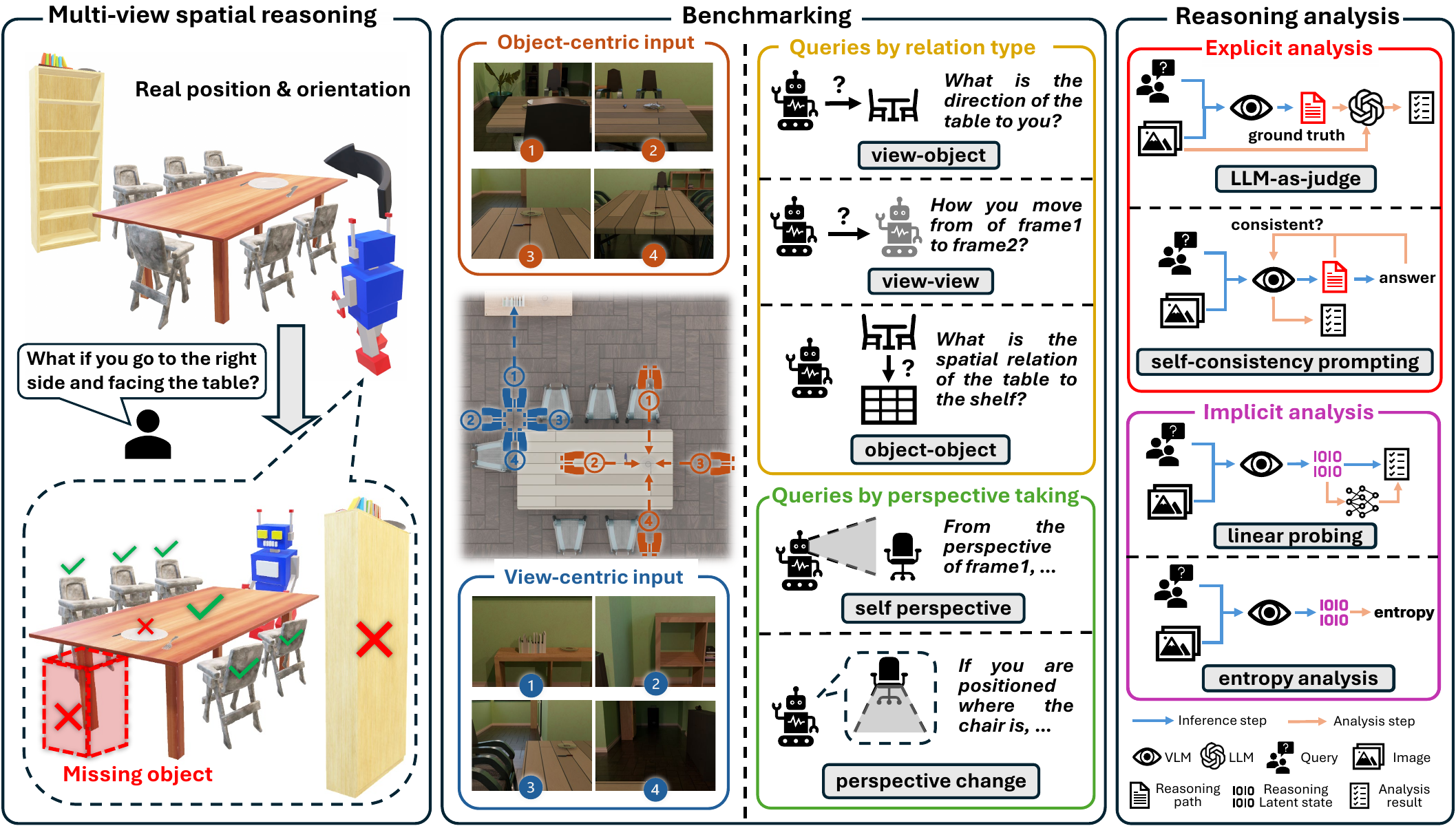} 
	\vspace{-0.1in}
	\caption{
		\textbf{Left:} Current VLMs struggle to maintain coherent spatial reasoning across multiple views (\checkmark~indicates consistent reasoning and \texttimes~denotes incorrect reasoning or localization).
		\textbf{Middle:} We assess multi-view spatial reasoning through fine-grained dimensions of diverse viewpoint spatial patterns and query types to capture key cognitive factors in spatial reasoning.
		\textbf{Right:} To further interpret VLM's successes and failures, we conduct explicit analysis of VLM's textual reasoning path and implicit analysis of VLM's latent token representation.
	}
	\vspace{-0.05in}
	\label{fig:fig1}
\end{figure*}

Although recent advances in VLMs have improved capabilities in multimodal perception, compositional reasoning, and task-oriented planning, they often rely on superficial cues, such as objects' colors, background layout, or statistics of 2D co-occurrences, to infer the spatial relations among objects, rather than forming coherent 3D representations across different views~\cite{zhang2025open3dvqa,yeh2025seeing,li2025viewspatial}. Consequently, as shown in Figure~\ref{fig:fig1} - left, their reasoning across different views often exhibits inconsistency, such as missing objects and incorrect cross-view spatial alignment, revealing a persistent gap between human-like spatial cognition and current VLM capabilities~\cite{stogiannidis2025mind,kamath2023s,jia2025omnispatial}.

This gap underscores the great need of fully understanding how VLMs construct, align, and maintain the spatial representations across different views. Existing benchmarks, however, are mostly limited to single-view images or video streams. 
Single images provide only partial context \cite{chen2024spatialvlm,ogezi2025spare,jia2025omnispatial,cheng2024spatialrgpt,zhang2025open3dvqa}, and miss occlusions, hidden spatial relationships and viewpoint-dependent cues across multiple perspectives \cite{jia2025omnispatial,cheng2024spatialrgpt}. A video stream could contain multiple frames of the same scene, but these frames usually do not correspond to well structured views that systematically probe the VLM's alignment, perspective taking and cross-view reasoning. Video streams also introduce temporal confounds such as motion tracking and temporal memory, which blur the focus of spatial reasoning \cite{ouyang2025spatial,cheng2025v,yang2025thinking,li2025viewspatial}. 

In contrast, a multi-view setting should present complementary views of a scene without temporal dependencies, and these views should carry structured geometric information that requires VLMs to maintain cross-view consistency, resolve occluded entities, and align spatial relations across perspectives. Existing multi-view benchmarks \cite{yeh2025seeing,yang2025mmsi,hong20233d} cannot meet these requirements, because they treat different views \emph{independently} and emphasize view coverage rather than cross-view coherence. Other multi-view benchmarks highlight spatial relationships across viewpoints \cite{gholami2025spatial,yin2025spatial,yeh2025seeing}, but lack fine-grained disentanglement of diverse factors that shape multi-view spatial reasoning.


At this end, our primary focus in this paper is two-fold. \emph{First}, we construct a new benchmark tailored for multi-view spatial reasoning, with complementary viewpoints that expose the underlying factors driving VLMs' successes and failures. \emph{Second}, based on this benchmark, we analyze VLM reasoning to uncover core challenges for achieving robust spatial inference in multi-view settings. In both of them, our primary approach is to draw on \emph{theories and findings from \textbf{cognitive science}}, ensuring that the spatial factors we model reflect human understanding of the physical world, and that our analytical framework aligns with human cognitive processes in spatial perception and reasoning.


\noindent \textbf{Benchmark.} 
We present \emph{ReMindView-Bench}, as the first cognitively grounded benchmark for multi-view spatial reasoning, with $>$50,000 spatial VQA pairs. ReMindView-Bench is designed to explicitly probe how VLMs construct, align and maintain spatial mental models across complementary viewpints. As shown in Figure~\ref{fig:fig1} - middle, ReMindView-Bench systematically organizes scenes according to object-centric and view-centric spatial patterns~\cite{tarr1989mental,biederman1987recognition}, and enables fine-grained diagnosis of cross-view reasoning and perspective taking over different query relationship types~\cite{gentner1983structure,hegarty2004mechanical,shepard1971mental,kosslyn1994image}. Guided by theories of human spatial cognition, ReMindView-Bench incorporates cognitive factors such as schema-based spatial memory, distance effects, and working memory constraints, which have not been fully evaluated in prior benchmarks~\cite{tversky2005functional,huttenlocher1991categories,loomis1992visual}.


Collectively, these dimensions form a structured cognitive evaluation protocol that isolates different sources of difficulty and provides fine-grained diagnostic insights into VLM spatial reasoning. Evaluations of 15 VLMs on ReMindView-Bench shows that both open-weight and closed-source models perform far below the human level, with accuracies plateauing $\sim$30–45\% compared to 81.5\% for humans. Across diverse settings, VLMs consistently perform better in object-centric than view-centric conditions, suffer pronounced degradation in cross-frame and perspective-changing reasoning, and lose robustness as the number of objects increases. These findings are the first systematic evidence of persistent weaknesses in geometric alignment and multi-view coherence, motivating deeper investigation into VLMs' spatial reasoning mechanisms.

\noindent\textbf{VLM Reasoning Analysis.} 
Our analysis is guided by cognitive theories of perceptual encoding, relational integration, and mental transformation~\cite{tarr1989mental,tversky2005functional,gentner1983structure}. Being different from the existing analysis that evaluates spatial reasoning as a single monolithic pass~\cite{trivedi2024self,shi2024judging}, we are the first to instruct VLMs to follow a multi-phase reasoning process that mirrors how humans construct and manipulate spatial representations across views, and then conduct \emph{fine-grained analysis on individual phases of this reasoning process}. As illustrated in Figure~\ref{fig:fig1} - right, our analysis includes both explicit and implicit manners. Explicitly, we assess the correctness of VLM's reasoning path using an LLM-as-a-judge framework~\cite{shi2024judging,trivedi2024self} in four reasoning phases, inspired by Johnson-Laird's theory of mental models~\cite{johnsonlaird1983mental,johnsonlaird1991deduction}. We further apply self-consistency prompting to evaluate whether VLM's generated rationales align with its internal decision-making~\cite{wang2022self,li2025drift,wang2025self}. Implicitly, we probe token-level latent representations using linear probing~\cite{stanczak2023latent,liu2024probing} to determine whether task-relevant information persists across reasoning phases, and analyze entropy dynamics~\cite{kossen2024semantic,ye2025uncertainty,xiong2024efficient} to quantify uncertainty and calibration as reasoning progresses.

These analyses link explicit reasoning paths with implicit latent representations in multi-view spatial reasoning. We find that VLMs remain reliable in early perceptual encoding but degrade sharply in later inferential stages, losing geometric coherence and cross-view consistency. Entropy trajectories reveal rising uncertainty and poor calibration in later phases. These results expose fundamental weakness in cross-view geometric alignment, stable inference progression, and confidence calibration, calling for cognitively grounded training strategies to improve spatial reasoning.

\section{Preliminaries}

\subsection{Spatial Reasoning in VLMs}

Spatial reasoning remains a core challenge for VLMs, as it requires grounding linguistic expressions into geometric and relational structures of the physical world rather than symbolic sequences. 
This explains why many VLMs are incapable of genuine spatial understanding but instead rely on language priors or dataset biases~\cite{hong20233d,yeh2025seeing,chen2024spatialvlm}. 

Achieving robust spatial reasoning would require human-like spatial perception and mental simulation~\cite{shepard1971mental,johnsonlaird1983mental}. Although spatial reasoning of VLMs has been enhanced by integrating explicit geometric priors \cite{chen2024spatialvlm,zheng2025learning}, 3D scene representations \cite{fan2025vlm, sun2025layoutvlm}, multi-view consistency learning \cite{hong20233d,xiang2022self} and spatial reasoning modules \cite{cheng2024spatialrgpt,ma2024spatialpin} to capture relational and perspective understanding beyond single-image semantics, current VLMs still lack 3D internal representations for flexible viewpoint transfer \cite{yang2025mmsi,yang2025thinking}. Cognitively grounded benchmarking and analysis are needed to align spatial reasoning tasks with mechanisms of human perception, construction, and mental transformation.

\subsection{Cognitive Perspective in Spatial Reasoning}
\label{sec:sec2.2}

Our benchmark design and analysis both build on the cognitive science perspective of humans' spatial understanding. 

\noindent\textbf{Benchmark design.} Research in visual cognition suggests to distinguish between \emph{object-centric} and \emph{view-centric} representations, which structure spatial attention and memory in different ways. Object-centric representation integrates multiple views of the same object, maintaining geometric identity across views~\cite{kahneman1992reviewing,scholl2001objects}. View-centric representation reflects spatial organization tied to the observer's position and viewing direction, producing context-dependent scene content~\cite{tarr1989mental,biederman1987recognition}. This dichotomy motivates our dual visual design in Figure \ref{fig:fig1}, where object-centric inputs test geometric integration across views and view-centric inputs evaluate perceptual stability under view changes.

We incorporate scene variability in layout diversity and object densities, to reflect the cognitive constraints and organizing principles in human spatial reasoning. According to the schema-based effect on spatial memory~\cite{tversky2005functional}, humans encode and recall spatial layouts based on high-level scene schemas (e.g., room type or functional zones) rather than absolute coordinates, and we hence incorporate diverse types of indoor rooms in our benchmark to examine how spatial reasoning generalizes across schema contexts. Further, the capacity limit of spatial working memory yields systematic declines in spatial reasoning as the number of objects increases \cite{clevenger2014working, wong2008visual}, and the distance effect \cite{huttenlocher1991categories,loomis1992visual} improves reasoning accuracy with larger spatial separations. Hence, our benchmark enforces fine-grained object–viewpoint distance control, and provides explicit labeling about the number of objects in each view.

Cognitive studies also revealed a hierarchical structure of spatial querying. Humans distinguish \textit{view–object}, \textit{view–view}, and \textit{object–object} relations~\cite{tversky2005functional,rieser1989access}, progressing from perceptual encoding to relational abstraction through recursive composition~\cite{gentner1983structure,hegarty2004mechanical}. In addition, perspective taking introduces additional cognitive complexity, engaging mechanisms of mental imagery~\cite{shepard1971mental,kosslyn1994image}. Accordingly, our benchmark differentiates queries between \textit{self-perspective} and \textit{perspective-change} with various types of spatial relations, to assess VLM's ability to simulate alternative views.


\begin{figure}[ht]
	\centering
	\vspace{-0.05in}
	\includegraphics[width=\linewidth]{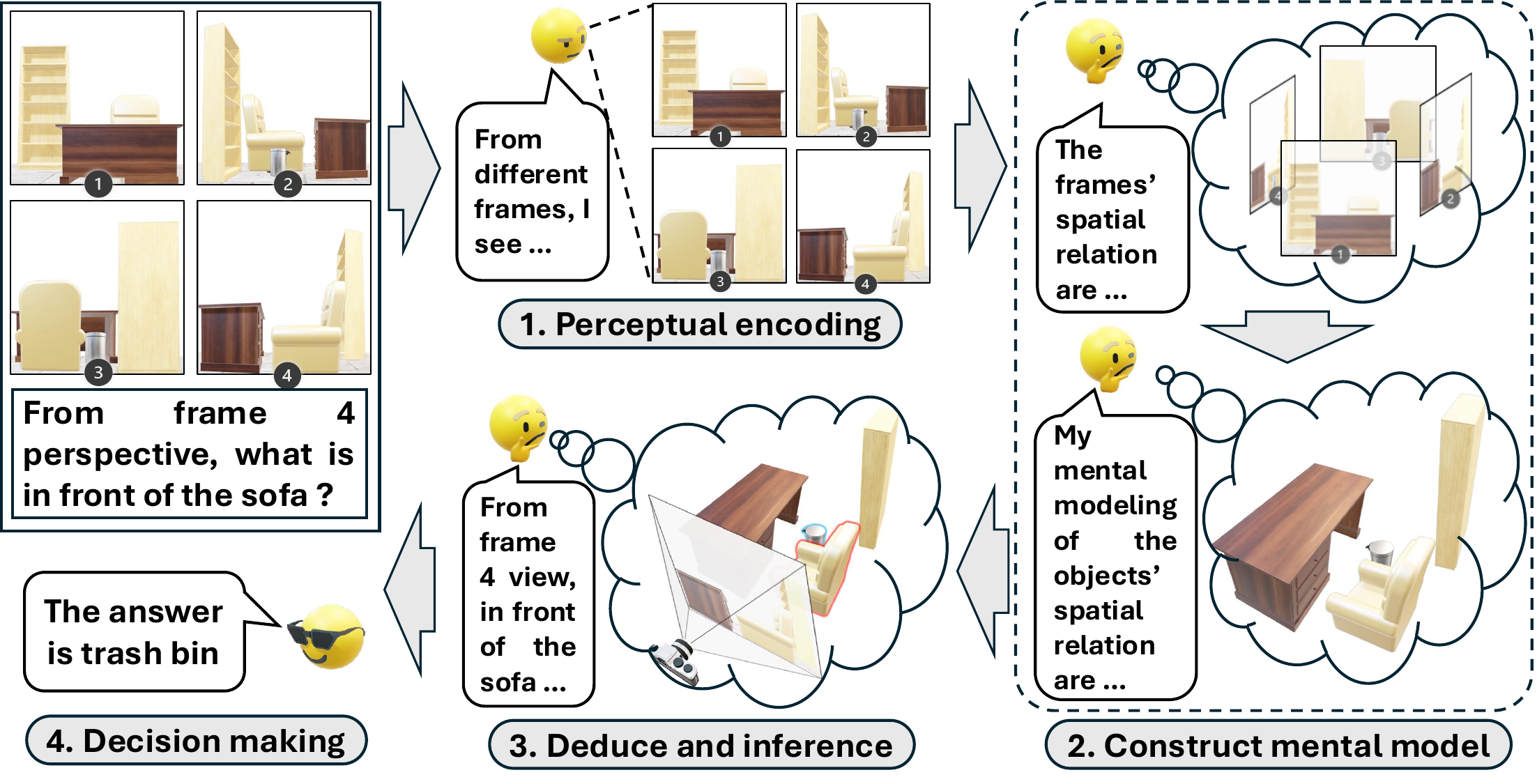}
	\vspace{-0.25in}
	\caption{Stages of humans' spatial mental modeling}
	\vspace{-0.1in}
	\label{fig:fig2}
\end{figure}

\begin{figure*}[ht]
	\vspace*{-2mm}
	\centering
	\vspace{-0.1in}
	\includegraphics[width=0.97\textwidth]{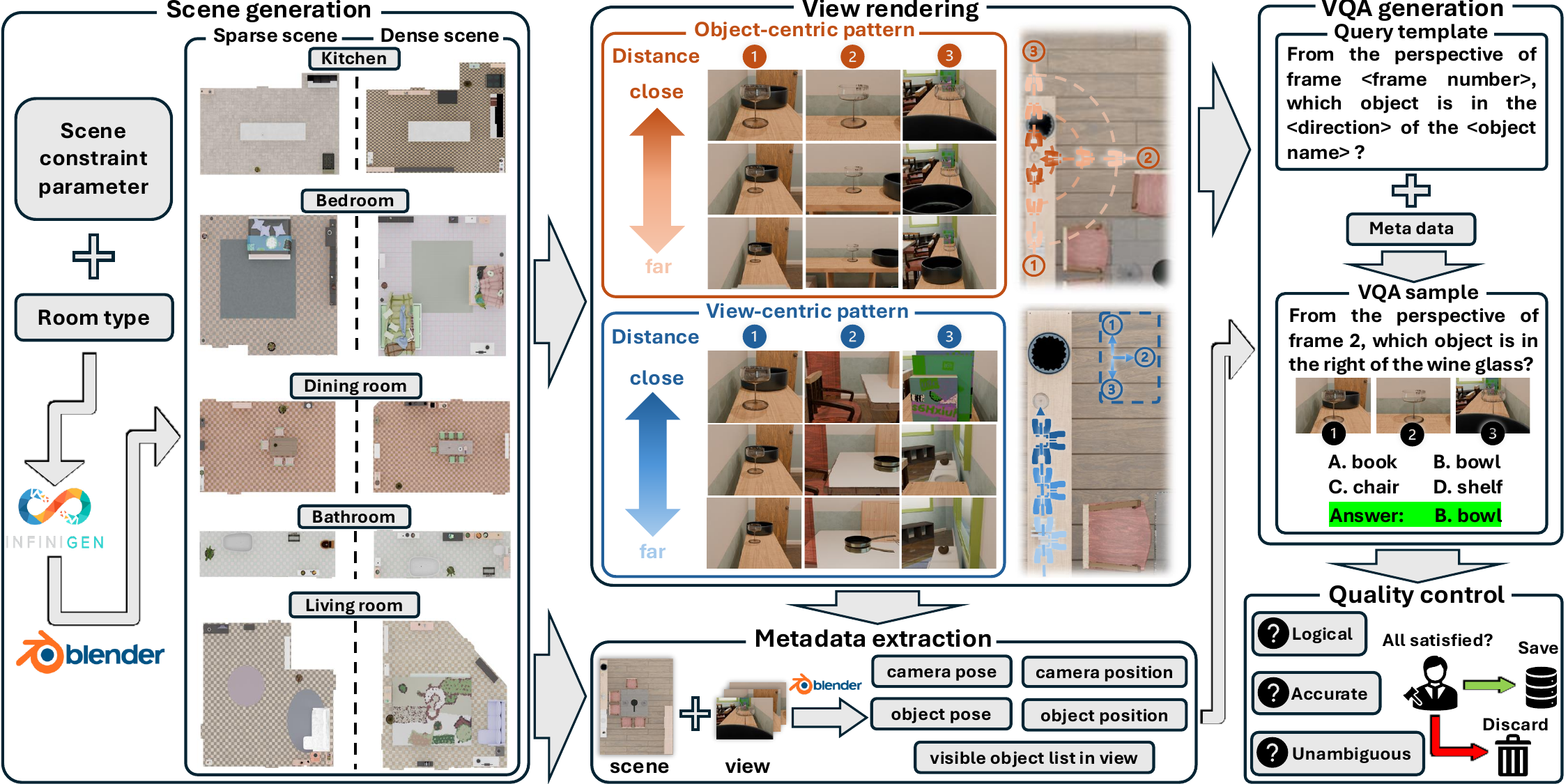}
	\vspace{-0.05in}
	\caption{\textbf{ReMindView-Bench construction pipeline}. It first generates diverse indoor scenes with various room types and object densities by adjusting scene constraint parameters in Infinigen. Next, multiple views are rendered in Blender with controlled camera–object distances and spatial patterns. VQA data is produced using predefined query templates, combined with metadata extracted from the scene and views.}
	\label{fig:fig3}
	\vspace{-0.15in}
\end{figure*}

\noindent\textbf{Analysis of spatial reasoning.} Cognitive science suggests that human spatial reasoning is not a monolithic pass, but a multi-stage progression as shown in Figure~\ref{fig:fig2}, including the initial encoding of perceptual input, the construction of coherent mental models that integrate information across multiple views, the deductive and inferential operations to address queries, and the final decision making ~\cite{johnsonlaird1983mental,johnsonlaird1991deduction}. This formulation is consistent with evidences from mental imagery and simulation~\cite{kosslyn1980image,kosslyn1994image}, mental rotation studies~\cite{shepard1971mental}, and spatial problem solving~\cite{hegarty2004mechanical}. 
Our analysis is aligned with these cognitive stages, and hence offers cognitively interpretable insights into where and how VLM reasoning diverges from human spatial cognition.

\subsection{Analysis Methods}

We will leverage both explicit and implicit analysis methods. Explicit analysis evaluates reasoning through generated textual rationales. LLM-as-a-judge ~\cite{trivedi2024self,shi2024judging} prompts LLMs to verify or criticize VLM's reasoning outputs, and self-consistency prompting~\cite{wang2022self,parcalabescu2023measuring,li2025drift,wang2025self} further improves robustness by aggregating multiple reasoning trajectories to assess whether rationales reflect internal decision making. However, when applied to spatial reasoning, textual explanations often under-specify geometry, overlook occlusion and perspective-taking, and can be spatially incorrect despite fluency, leading to inflated assessments. 

Implicit analysis, instead, examines latent representations to uncover the internal reasoning dynamics. Probing studies~\cite{zhang2025reasoning,stanczak2023latent,liu2024probing} test whether task-relevant structure is linearly recoverable from activations, and entropy-based analysis~\cite{kossen2024semantic,ye2025uncertainty,xiong2024efficient} tracks the propagation of uncertainty to reveal the decision stability. Yet for spatial reasoning, latent activations often entangle geometry with visual texture, and entropy changes signal uncertainty without localizing the cause of failure, limiting interpretability.

We integrate both manners. Explicit analysis elucidates how VLMs articulate the reasoning process, and implicit analysis reveals internal representations underlying these articulations. Combining them offers fine-grained understandings of how VLMs construct and maintain spatial representations over reasoning phases, as shown in Figure \ref{fig:fig2}.

\vspace{-0.2in}
\section{Benchmark}
\vspace{-0.05in}
\label{sec:benchmark}
ReMindView-Bench 
consists of $>$50,000 multi-choice VQA pairs derived from 100 procedurally generated indoor scenes, and is constructed through a fully automated and physically grounded pipeline. 

\vspace{-0.05in}
\subsection{Scene Variability}
\label{subsec:scene_variability}
\vspace{-0.05in}
ReMindView-Bench enforces two primitive forms of human spatial cognition, namely relative direction and relative distance~\cite{levinson2003space,tversky2005functional}. 
Based on the cognitive factors affecting humans' spatial reasoning in Section~\ref{sec:sec2.2}, it provides a fine-grained collection of VQA tasks from various cognitive science perspectives, as shown in Figure \ref{fig:fig3}. For the vision data of VQA tasks, its variability spans the following 4 factors:

\begin{enumerate}
	\item \textit{Room type}, including dining room, kitchen, bathroom, living room, and bedroom.
	\item \textit{Viewpoint spatial pattern}, including view-centric and object-centric configurations.
	\item  \textit{Level of distance}, including fine-grained levels of distances between the camera and indoor objects, which control the scene's spatial proximity.
	\item \textit{Number of visible objects} across all rendered views.	
\end{enumerate}

The varability of query data spans another 4 factors:
\begin{enumerate}
	\item \textit{Query type}, as whether the question concerns relative distance or relative direction.
	\item \textit{Relation type}, defined as the relational structure among viewpoints and objects being queried, including view–object, object–object, and view–view relations.
	\item \textit{Perspective taking}, as whether spatial reasoning is grounded in the current camera viewpoint or shifted to another object-centered viewpoint.
	\item \textit{Cross-frame reasoning}, as whether all entities involved in the query are co-observed within a single frame, or instead require multi-view integration.
\end{enumerate}


We provide more details about the VQA classification scheme in Appendix \ref{appendix:B1}. Some examples of VQA queries are shown in Figure \ref{fig:fig1}, and we provided a richer set of VQA task examples in Appendix \ref{appendix:B2}. Statistics of the ReMindView-Bench benchmark are in Appendix \ref{appendix:B3}.

\vspace{-0.05in}
\subsection{Construction Pipeline}
\vspace{-0.05in}
\label{subsec:construction_pipeline}
As shown in Figure~\ref{fig:fig3}, the construction pipeline of ReMindView-Bench integrates procedural 3D scene synthesis, viewpoint-controlled rendering, structured metadata extraction, and template-based question generation. 

\noindent\textbf{Scene generation.} We procedurally generate diverse indoor scenes using Infinigen~\cite{raistrick2024infinigen}, with controllable parameters for room type, object density, and spatial layout. 
Detailed constraint settings are in Appendix \ref{appendix:C1}. This controlled generation allows systematic variation in clutter and spatial arrangement, challenging VLMs across various perceptual and geometric conditions. Each 3D scene is imported into Blender~\cite{blender} for rendering and metadata computation.

\noindent\textbf{View rendering.} For each scene, we render 4 canonically orthogonal views for each target object, following both object-centric and view-centric spatial patterns. Each pattern is captured at 10 discrete object–viewpoint distances, precisely regulating camera proximity and providing multi-scale spatial cues. Formal definitions of distance levels and strategy for viewpoint selection are in Appendix \ref{appendix:C2}. These settings yield structured yet diverse visual observations that capture core elements of spatial cognition.

\noindent\textbf{Metadata extraction.} Each scene–view pair is annotated with geometric metadata exported from Blender, including camera pose, object pose, and list of visible objects. It supports accurate construction of relational queries and provides geometric grounding for VQA generation.

\begin{figure}[t] \centering \includegraphics[width=\linewidth]{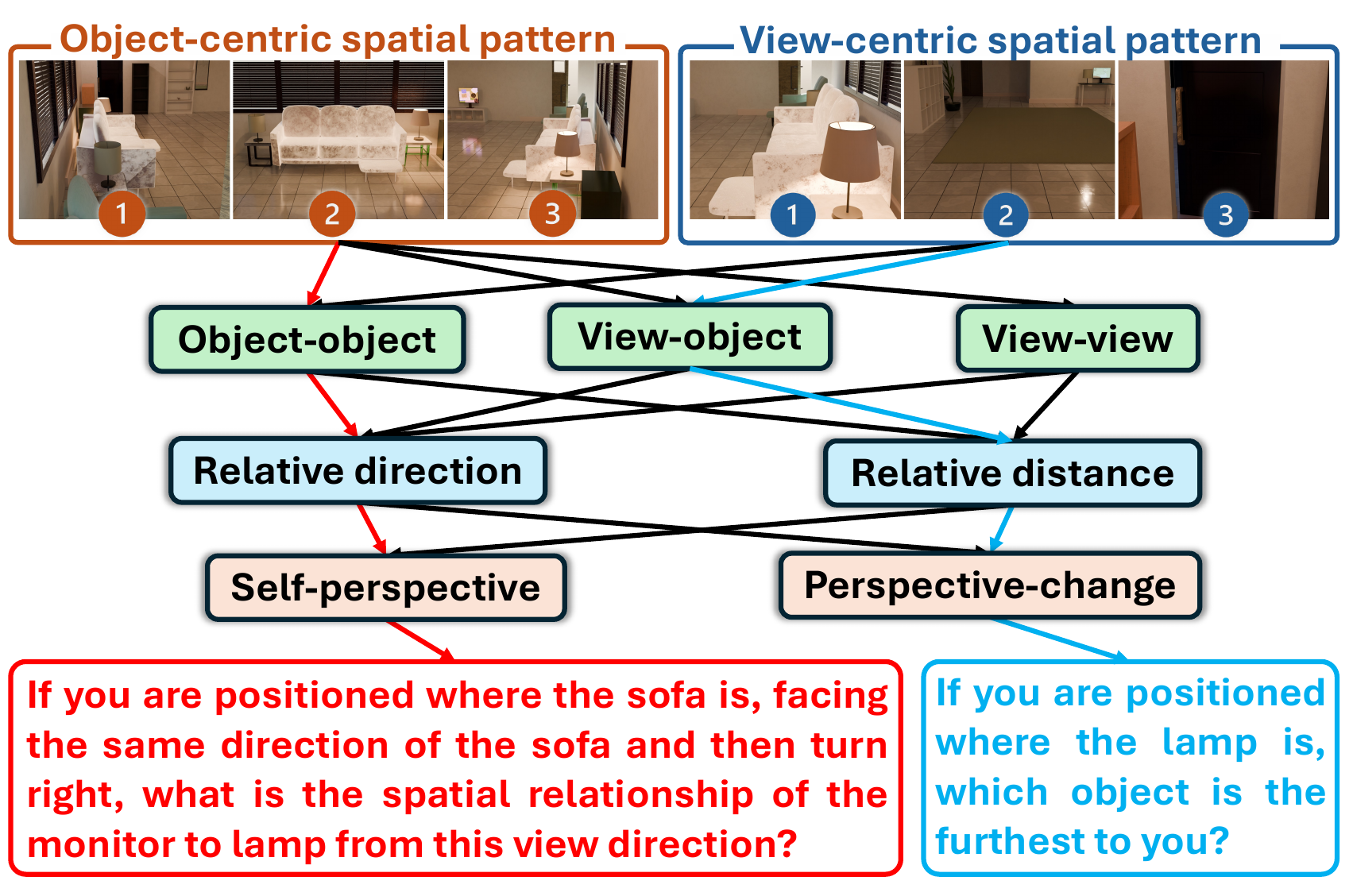} \vspace{-0.25in}\caption{Query-related label combinations to generate VQA pairs} \label{fig:VQ_generation} \vspace{-0.2in} \end{figure}

\noindent\textbf{VQA generation.} Using the extracted metadata, we instantiate 22 predefined query templates covering relative direction, relative distance, relation type and perspective-taking as shown in Figure~\ref{fig:VQ_generation}. For each query, candidate answers are drawn from the visible objects in the scene, and the correct answer is computed from the metadata. An example is shown in Figure~\ref{fig:fig3}, with full query instantiation and answer derivation procedures detailed in Appendix \ref{appendix:C3}. 


\newcommand{\up}{\textsuperscript{$\uparrow$}}
\newcommand{\down}{\textsuperscript{$\downarrow$}}

\definecolor{rowgray}{RGB}{247,247,247}
\definecolor{c1}{gray}{0.66}
\definecolor{c2}{gray}{0.78}
\definecolor{c3}{gray}{0.9}
\newcommand{\bestcell}{\cellcolor{c1}}
\newcommand{\secondcell}{\cellcolor{c2}}
\newcommand{\thirdcell}{\cellcolor{c3}}
\renewcommand{\arraystretch}{0.9} 
\newcolumntype{N}{>{\footnotesize\arraybackslash}c}  

\begin{table*}[!t]
	\centering
	\renewcommand{\arraystretch}{0.9} 
	\small
	\setlength{\tabcolsep}{1 pt} 
	\renewcommand\theadfont{\normalsize\bfseries}
	\begin{tabular}{
			l !{\vrule width 0.8pt} c !{\vrule width 0.8pt}
			*{10}{N} !{\vrule width 0.8pt}
			*{9}{N}
			*{7}{c} c c c !{\vrule width 0.8pt} 
			*{6}{c} c c c 
		}
		\toprule
		\multirow{2}{*}{\textbf{Methods}} &
		\multirow{2}{*}{\textbf{Avg.}} &
		\multicolumn{10}{c!{\vrule width 0.8pt}}{\textbf{Object-centric Viewpoint Spatial Pattern}} &
		\multicolumn{9}{c}{\textbf{View-centric Viewpoint Spatial Pattern}} \\
		\cmidrule(lr){3-12}\cmidrule(lr){13-21}
		& &
		\textbf{V–V} & \textbf{V–O} & \textbf{O–O} & \textbf{SP} & \textbf{PC} & \textbf{Dir.} & \textbf{Dis.} & \textbf{CF} & \textbf{NCF} & \textbf{Avg.} &
		\textbf{V–O} & \textbf{O–O} & \textbf{SP} & \textbf{PC} & \textbf{Dir.} & \textbf{Dis.} & \textbf{CF} & \textbf{NCF} & \textbf{Avg.} \\
		\midrule
		Random Choice & 30.9 & 31.2 & 31.1 & 30.1 & 31.6 & 31.2 & 30.9 & 31.9 & 32.2 & 25.8 & 31.2 & 31.0 &  30.4 & 30.8 & 31.1 & 30.5 & 31.2 & 31.5 & 25.5 & 30.7 \\
		\addlinespace[2pt]
		\rowcolor{rowgray}\multicolumn{21}{l}{\textit{ReMindView-Bench (small) Perf.}}\\
		{\dag}Human Level & 81.5 & 84.2 & 86.8 & 79.9 & 87.4 & 76.1 & 80.3 & 88.7 & 76.3 & 89.6 & 82.5 & 81.1 & 78.6 & 83.4 & 75.9 & 74.3 & 86.1 & 74.6 & 88.9 & 80.9 \\
		{\dag}Gemini–2.5 Pro & 43.2 & 41.1 & 45.3 & 46.5 & 48.7 & 42.9 & 41.0 & 47.8 & 40.8 & 36.2 & 44.8 & 43.6 & 39.5 & 43.2 & 38.9 & 34.4 & 55.2 & 38.4 & 29.1 & 42.3 \\
		\addlinespace[2pt]
		\rowcolor{rowgray}\multicolumn{21}{l}{\textit{Proprietary Models}}\\
		GPT-4o & \secondcell 38.6\up & \thirdcell 37.8\up & \thirdcell 41.2\up & \secondcell 41.9\up & \thirdcell 44.1\up & \secondcell 38.9\up & \secondcell 37.4\up & \secondcell 43.2\up & \secondcell 41.8\up & \secondcell 32.3\up & \secondcell 40.6\up & \secondcell 39.8\up & \secondcell 34.7\up & \thirdcell 39.3\up & \secondcell 34.9\up & 29.9\down & \secondcell 51.6\up & \secondcell 40.2\up & \thirdcell 31.5\up & \secondcell 37.0\up \\
		Claude–4–Sonnet & 34.1\up & 28.9\down & 39.9\up & 32.5\up & 42.4\up & 31.1\down & 31.2\up & 37.8\up & 35.6\up & 27.4\up & 33.8\up & 37.2\up & 31.9\up & 38.7\up & 31.2\down & 28.2\down & 47.7\up & 35.4\up & 29.8\up & 34.6\up \\
		Gemini–2.5 Pro & \bestcell 43.1\up & \bestcell 41.5\up & \bestcell 44.3\up & \bestcell 45.7\up & \bestcell 47.9\up & \bestcell 42.6\up & \bestcell 41.1\up & \bestcell 46.3\up & \bestcell 45.8\up & \bestcell 41.2\up & \bestcell 44.5\up & \bestcell 43.7\up & \bestcell 39.2\up & \bestcell 42.8\up & \bestcell 39.9\up & \bestcell 35.4\up & \bestcell 54.8\up & \bestcell 43.4\up & \bestcell 34.1\up & \bestcell 41.6\up \\
		\addlinespace[2pt]
		\rowcolor{rowgray}\multicolumn{21}{l}{\textit{Open-source Models}}\\
		Qwen2.5–VL–3B & 32.2\up & 32.5\up & 33.3\up & 30.2\up & 36.5\up & 28.2\down & 29.3\down & 36.5\up & 33.8\up & 26.2\up & 32.2\up & 34.9\up & 29.7\down & 34.8\up & 30.3\down & \thirdcell 30.8\up & 35.7\up & 32.4\up & 31.6\up & 32.3\up \\
		Qwen2.5–VL–7B & 34.2\up & 36.8\up & 36.6\up & 32.1\up & 39.9\up & 29.8\down & 30.6\down & 42.5\up & 35.2\up & \thirdcell 32.6\up & 35.2\up & 34.3\up & 31.1\up & 35.9\up & 29.9\down & 29.1\down & 39.9\up & 34.3\up & \secondcell 32.6\up & 32.6\up \\
		Qwen2.5–VL–32B & \thirdcell 37.5\up & \secondcell 37.9\up & \secondcell 41.8\up & \thirdcell 35.4\up & \secondcell 44.8\up & \thirdcell 33.5\up & \thirdcell 35.4\up & \thirdcell 43.1\up & \thirdcell 39.2\up & 31.7\up & \thirdcell 38.4\up & \thirdcell 38.3\up & \thirdcell 34.3\up & \secondcell 39.5\up & \thirdcell 33.8\up & \secondcell 32.1\up & \thirdcell 45.0\up & \thirdcell 40.1\up & 32.1\up & \thirdcell 36.4\up \\
		InternVL3.5–2B & 31.6\up & 39.7\up & 32.5\up & 29.1\down & 33.4\up & 28.7\down & 29.9\down & 40.0\up & 35.6\up & 27.1\up & 33.8\up & 28.9\down & 28.0\down & 30.7\down & 26.6\down & 26.2\down & 33.1\up & 29.1\down & 24.5\down & 28.5\down \\
		InternVL3.5–8B & 25.3\down & 25.9\down & 27.3\down & 24.5\down & 30.3\down & 22.3\down & 22.5\down & 31.3\up & 25.3\down & 18.8\down & 26.0\down & 26.1\down & 22.4\down & 26.8\down & 22.2\down & 21.6\down & 29.8\down & 27.6\down & 19.8\down & 24.3\down \\
		InternVL3.5–38B & 32.6\up & 33.3\up & 34.3\up & 30.1\down & 36.6\up & 29.2\down & 28.5\down & 39.6\up & 37.0\up & 28.5\up & 32.8\up & 34.3\up & 30.6\up & 35.2\up & 30.2\down & 30.1\down & 37.4\up & 33.0\up & 30.3\up & 32.4\up \\
		LLaVA–Video–7B & 32.4\up & 33.1\up & 34.5\up & 31.2\up & 36.1\up & 28.9\down & 30.4\down & 38.3\up & 33.5\up & 30.4\up & 33.2\up & 33.2\up & 30.6\up & 34.1\up & 29.4\down & 28.1\down & 39.2\up & 34.1\up & 31.8\up & 32.4\up \\
		LLaVA–OneVision–0.5B & 22.3\down & 23.1\down & 23.8\down & 21.4\down & 25.7\down & 20.5\down & 21.3\down & 27.1\down & 23.8\down & 19.5\down & 23.3\down & 22.8\down & 20.9\down & 23.6\down & 20.3\down & 19.7\down & 26.4\down & 22.7\down & 20.8\down & 22.3\down \\
		LLaVA–OneVision–7B & 33.8\up & 35.0\up & 36.4\up & 32.7\up & 38.9\up & 30.5\down & 31.8\down & 40.1\up & 34.6\up & 32.5\up & 35.0\up & 34.6\up & 31.8\up & 35.9\up & 30.9\down & 29.2\down & 41.3\up & 35.9\up & 33.2\up & 33.9\up \\
		\addlinespace[2pt]
		\rowcolor{rowgray}\multicolumn{21}{l}{\textit{Spatial Models}}\\
		SpatialVLM & 31.6\up & 38.9\up & 30.3\down & 28.8\down & 32.5\up & 27.1\down & 29.3\down & 38.1\up & 34.7\up & 25.2\down & 32.6\up & 31.6\up & 28.4\down & 33.1\up & 27.5\down & 27.9\down & 34.3\up & 31.0\up & 24.6\down & 30.1\down \\
		SpatialMLLM & 33.1\up & 35.3\up & 32.9\up & 31.9\up & 36.2\up & 29.3\down & 31.1\down & 37.1\up & 33.6\up & 32.2\up & 33.4\up & 34.9\up & 29.8\down & 34.6\up & 30.5\down & 30.4\down & 36.3\up & 35.0\up & 27.6\up & 32.3\up \\
		SpaceQwen2.5-VL & 28.4\down & 32.5\up & 29.3\down & 27.1\down & 29.3\down & 27.3\down & 27.6\down & 31.6\up & 30.5\down & 24.9\down & 29.1\down & 27.5\down & 27.6\down & 28.4\down & 26.9\down & 26.5\down & 29.8\down & 28.3\down & 23.2\down & 27.6\down \\
		\bottomrule
	\end{tabular}
	\vspace{-0.05in}
	\caption{\textbf{Evaluation results on \textit{ReMindView-Bench}.} \begin{tabular}{@{}l@{}}
			\cellcolor{c1}\textbf{Dark gray}
		\end{tabular}, 
		\begin{tabular}{@{}l@{}}
			\cellcolor{c2}\textbf{gray}
		\end{tabular}, 
		\begin{tabular}{@{}l@{}}
			\cellcolor{c3}\textbf{light gray}
		\end{tabular} indicate the best, second best and third best result among all models, $\uparrow$ and $\downarrow$ indicate performance higher or lower than the random-guess baseline. $\dagger$ indicates models evaluated on the \textit{ReMindView-Bench (small)} subset. \textbf{V–V}, \textbf{V–O}, and \textbf{O–O} denote \textit{view–view}, \textit{view–object}, and \textit{object–object} relations; \textbf{SP} and \textbf{PC} represent \textit{self-perspective} and \textit{perspective-changing}; \textbf{Dir.} and \textbf{Dis.} correspond to \textit{relative direction} and \textit{relative distance} query type; \textbf{CF} and \textbf{NCF} represent \textit{cross-frame} and \textit{non-cross-frame} reasoning tasks.}
	\label{tab:eval}
	\vspace{-0.15in}
\end{table*}

\section{Evaluation on ReMindView-Bench}

We evaluated current VLMs' performance on multi-view spatial reasoning, using ReMindView-Bench benchmark.

\subsection{Evaluation Setup}

\textbf{Benchmark Models.} We evaluate 15 VLMs that take multi-image input. 
 Proprietary models include \textit{Gemini-2.5-Pro} \cite{comanici2025gemini}, \textit{GPT-4o} \cite{openai2024gpt4o} and \textit{Claude-4-Sonnet} \cite{anthropic2024claude4sonnet}. Open-source models include \textit{InternVL3.5} \cite{chen2024expanding}, \textit{Qwen2.5-VL} \cite{bai2025qwen2}, \textit{LLaVA-Video} \cite{lin2024video} and \textit{LLaVA-OneVision} \cite{li2024llava} families with different parameter sizes. We also included models that are explicitly designed for spatial reasoning, namely \textit{SpatialVLM} \cite{chen2024spatialvlm}, \textit{SpatialMLLM} \cite{wu2025spatial} and \textit{SpaceQwen2.5-VL} \cite{jia2025omnispatial}. Evaluations use a zero-shot setting with a unified prompting strategy, and details are in Appendix \ref{appendix:D}.

\noindent\textbf{Metric Design and Baselines.} We consider spatial reasoning tasks as multi-choice questions (MCQs), and use the task accuracy based on exact matching as the primary evaluation metric. We provide a random-guess baseline with the random selection accuracy. We further sample a subset of 660 VQA samples with 30 VQAs per query template, referred to as \textit{ReMindView-Bench (small)}, for human evaluators to independently answer each question and evaluate the performance. For comparison, we also report Gemini–2.5 Pro's performance on \textit{ReMindView-Bench (small)}.

\subsection{Main Results}
Our evaluations reveal a pronounced gap between human spatial reasoning and current VLMs' capabilities.


\noindent\textbf{Object-centric reasoning remains easier than view-centric reasoning.}
Table \ref{tab:eval} shows that nearly all VLMs performed better in object-centric spatial patterns, indicating that VLMs rely on within-view visual cues but fail to maintain relational consistency across view. Particularly, tasks involving perspective-changing or relative direction queries reveal large performance degradation, underscoring the difficulty of aligning spatial geometry and insufficient robustness to viewpoint transformations in spatial reasoning.

\noindent\textbf{Viewpoint transformations amplify the reasoning instability.}
Table \ref{tab:eval} shows that VLMs have a performance drop $>$4\% in PC settings, compared to SP settings. Models that perform well under fixed viewpoints fail to preserve spatial consistency when reasoning involves perspective changing, indicating weak cross-view constancy and reference-frame alignment. From a cognitive perspective, this deficiency mirrors early-stage perceptual reasoning without mature allocentric transformation~\cite{tversky2005functional,loomis1992visual}, and shows that current attention mechanisms fail to encode cross-view correspondences, leading to disrupted spatial continuity and limited generalization across geometric variations.

\begin{figure}[ht] \centering \vspace{-0.05in}\includegraphics[width=\linewidth]{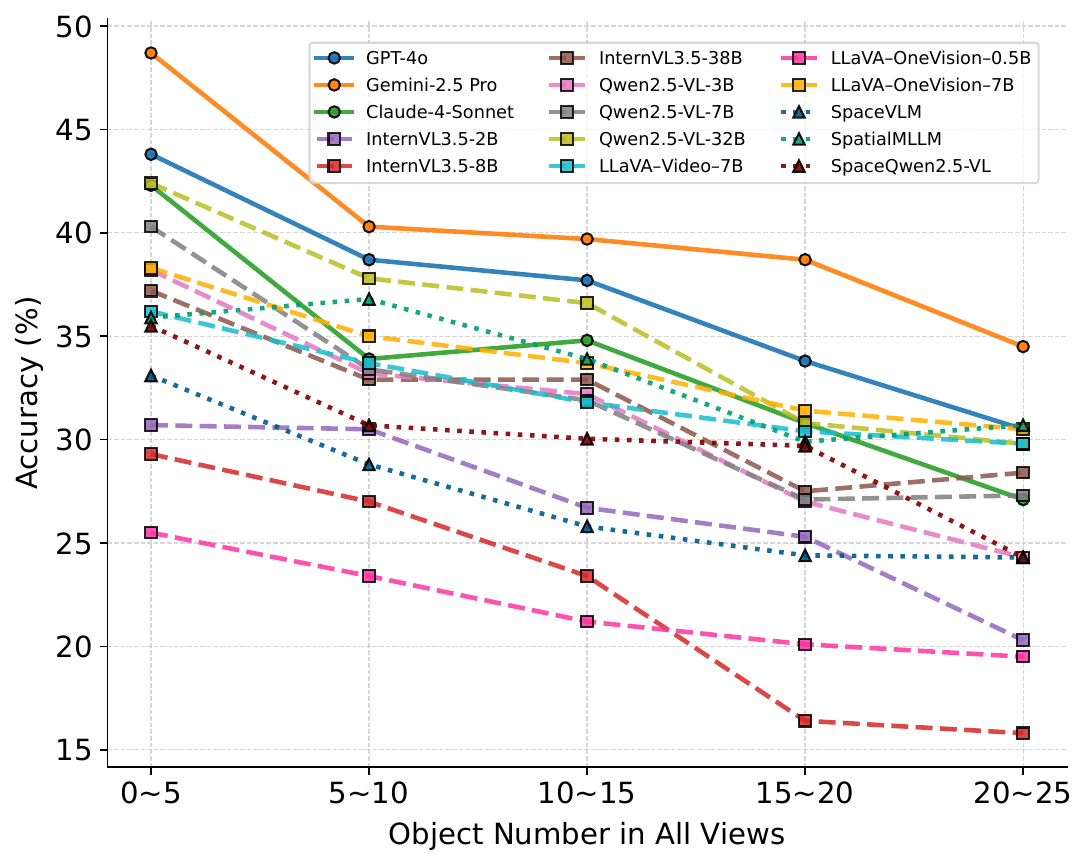} \vspace{-0.3in}\caption{Task accuracy with with different numbers of objects}\label{fig:object_number} \vspace{-0.25in} \end{figure}

\begin{figure}[ht] \centering \vspace{-0.05in}\includegraphics[width=\linewidth]{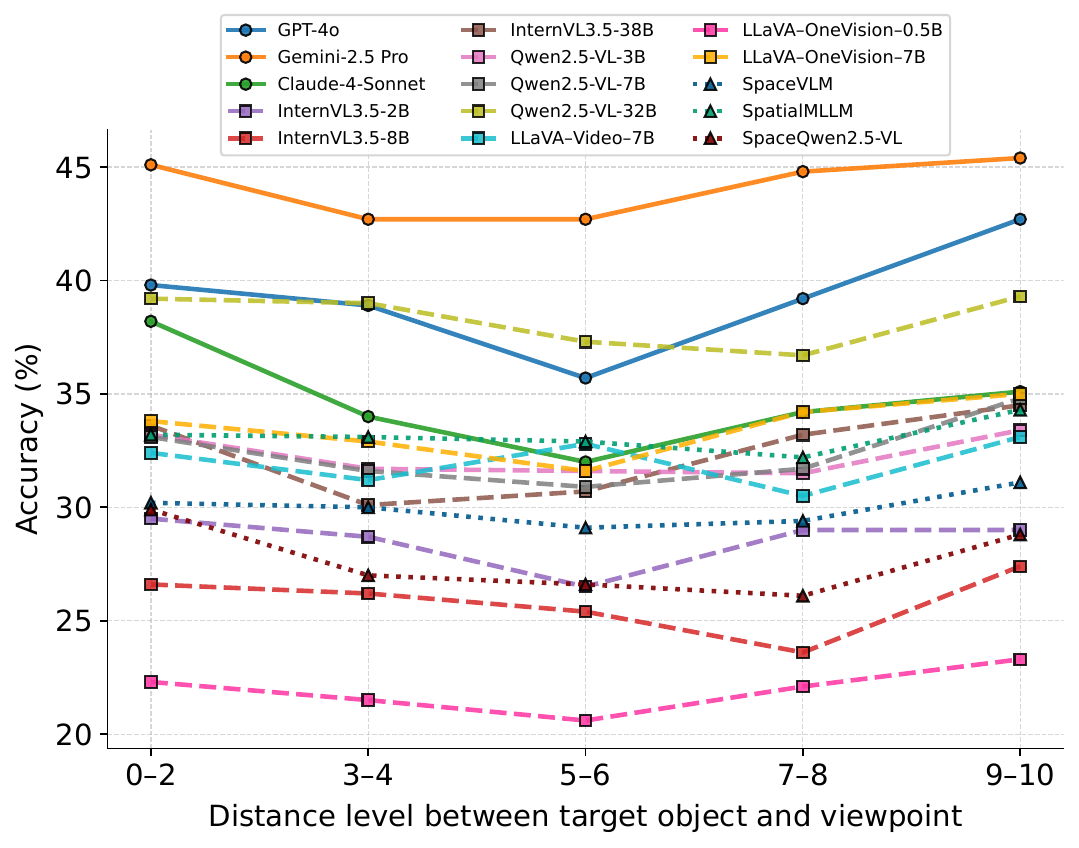} \vspace{-0.25in}\caption{Task accuracy with varying object-viewpoint distances} \label{fig:distance} \vspace{-0.2in} \end{figure}

\noindent\textbf{Cross-frame reasoning is more challenging than within-frame reasoning.}
As shown in Table~\ref{tab:eval}, all VLMs exhibit a clear performance drop when reasoning requires integrating information across multiple frames. This pattern holds consistently under both object-centric and view-centric spatial patterns, highlighting the difficulty of maintaining coherent spatial representations over disjoint visual inputs. The large CF to NCF gap indicates that current VLMs fail to geometrically align scene elements across complementary views, relying instead on static perceptual features in single frames. These findings show that cross-frame reasoning, which requires integrating spatial correspondences and relations, is the key to improve multi-view spatial cognition.

\noindent\textbf{Number of objects and object-viewpoint distance jointly constrain spatial reasoning.}
As shown in Figure~\ref{fig:object_number}, increasing the number of objects in the scene leads to accuracy degradation, which suggests that cluttered scenes amplify relational ambiguity and visual occlusion, hindering VLMs from maintaining coherent object–relation mappings. Figure~\ref{fig:distance} shows that the accuracy across different object-viewpoint distances is comparatively stable. A shallow U-shaped trend is observed, where close and far distances yield higher accuracy and mid-range distances have degraded performance. This pattern suggests that extreme distances provide clearer geometric cues, either through strong visual overlap or pronounced parallax. 
These results demonstrate that current VLMs lack scalable mechanisms for spatial compositionality, struggling both with object-level clutter and geometric variance across views.

\section{Explicit VLM Reasoning Analysis}
Explicit analysis examines the correctness of a VLM's reasoning phases and the consistency between reasoning and decision, pinpointing where spatial reasoning succeeds or breaks across reasoning phases.

\begin{figure}[ht] \vspace{-0.05in} \centering \includegraphics[width=\linewidth]{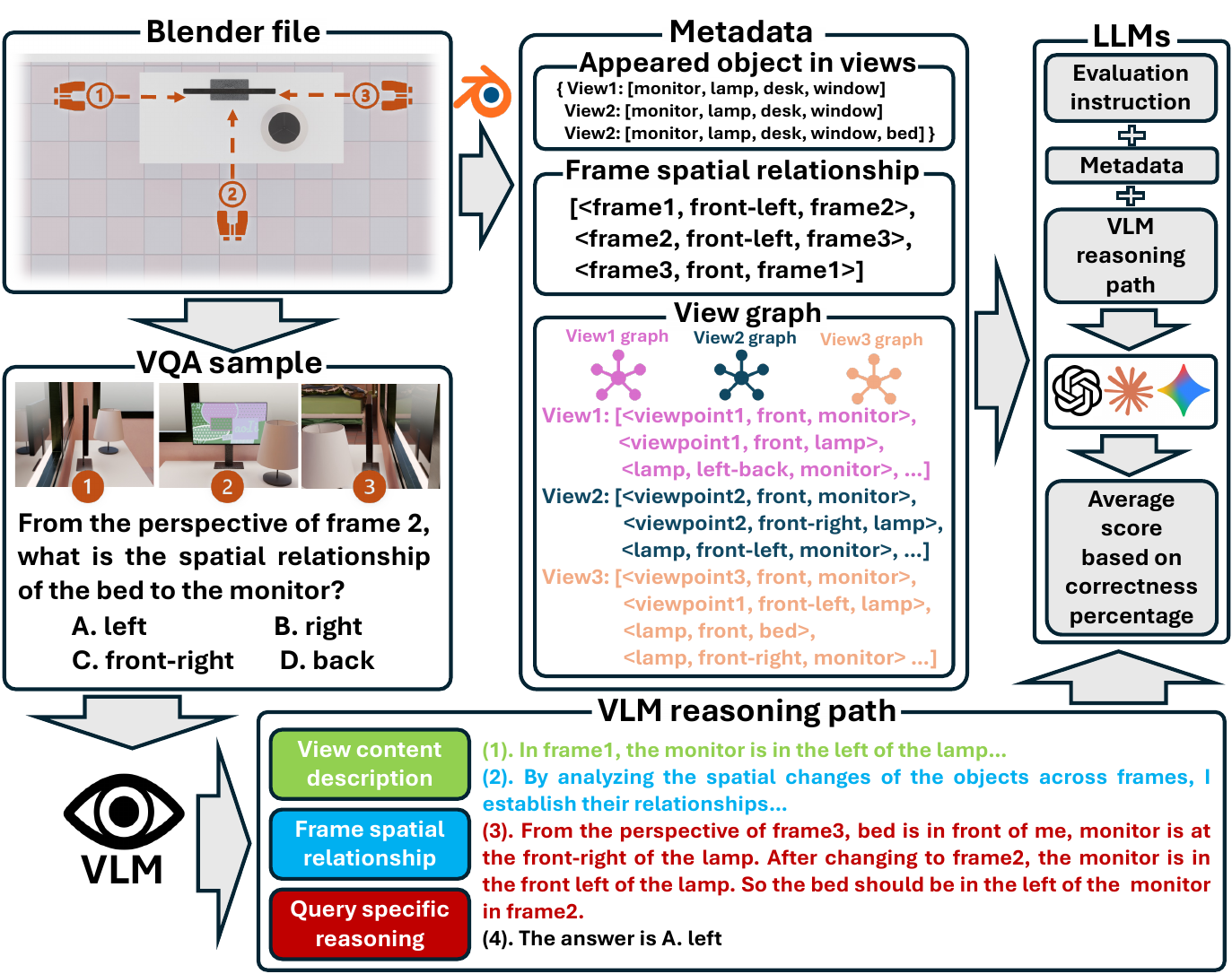} \vspace{-0.2in}\caption{The LLM-as-a-judge workflow for evaluating the correctness of VLM's spatial reasoning path} \label{fig:LLM_as_judge} \vspace{-3mm} \end{figure}

\subsection{LLMs-as-a-judge Evaluation}
As shown in Figure~\ref{fig:LLM_as_judge}, we use LLMs as judges to evaluate the correctness of VLM's spatial reasoning path, through a lens grounded in human spatial cognition. The analysis begins with extracting rich metadata from the rendered scenes in Blender, including (i) the list of objects in each view, (ii) spatial relationships among viewpoints and (iii) spatial relationships between objects and viewpoints, and organzing such metadata as separate view graphs. In each graph, nodes represent objects or viewpoints and edges encode spatial relations, ensuring that all visible entities across views are represented within a unified relational structure. 

We follow the cognitive science framework in Section~\ref{sec:sec2.2} to instruct VLM's reasoning with the 4-phase template: (1) description of view content spatial relationship, (2) alignment of frame-level spatial relationship, (3) query-specific spatial reasoning, and (4) final decision. A panel of LLMs receive VLM's reasoning path, ground-truth metadata and evaluation instruction, to calculate a phase-wise score of reasoning correctness based on the consistency between spatial relationship in view graphs and VLM's reasoning path by step, and scores from all LLM judges are averaged. More details are in Appendix \ref{appendix:E1}.

\begin{table}[t]
	\centering
	\vspace{2mm}
	\resizebox{\linewidth}{!}{
			{\fontsize{8}{9}\selectfont
		\begin{tabular}{lcccccc}
			\toprule
			\multirow{2}{*}{\textbf{Model}} & 
			\multicolumn{3}{c}{\textbf{Correct (\%)}} & 
			\multicolumn{3}{c}{\textbf{Incorrect (\%)}} \\
			\cmidrule(lr){2-4} \cmidrule(lr){5-7}
			& \textbf{1} & \textbf{2} & \textbf{3} 
			& \textbf{1} & \textbf{2} & \textbf{3} \\
			\midrule
			Qwen2.5-VL-3B  & 73.2 & 43.8 & 32.6 & 68.7 & 31.4 & 18.5 \\
			Qwen2.5-VL-7B  & 76.1 & 58.2 & 36.3 & 72.5 & 43.6 & 21.9 \\
			Qwen2.5-VL-32B & 81.9 & 63.5 & 38.8 & 76.8 & 44.2 & 24.6 \\
			\bottomrule
		\end{tabular}}
	}
	\vspace{-0.05in}
	\caption{Phase-wise reasoning correctness of Qwen2.5-VL family}
	\vspace{-0.1in}
	\label{tab:llm_judge}
\end{table}

\begin{table}[t]
	\centering
	\small
	\setlength{\tabcolsep}{5pt}
		{\fontsize{8}{9}\selectfont
	\begin{tabular}{lccc}
		\toprule
		\textbf{Model} & \textbf{Correct (\%)} & \textbf{Incorrect (\%)} & \textbf{Overall (\%)} \\
		\midrule
		Qwen2.5-VL-3B  & 64.8 & 32.5 & 48.6 \\
		Qwen2.5-VL-7B  & 74.1 & 41.7 & 59.3 \\
		Qwen2.5-VL-32B & 83.6 & 54.2 & 68.4 \\
		\bottomrule
	\end{tabular}}
	\vspace{-0.05in}
	\caption{Self-consistency prompting of Qwen2.5-VL family}
	\vspace{-0.2in}
	\label{tab:self_consistency}
\end{table}

\noindent\textbf{Phase-wise reasoning reveals strong in-frame performance but degraded cross-frame inference.}
As shown in Table~\ref{tab:llm_judge}, the LLM-as-a-judge evaluation exposes a phase-dependent degradation pattern. The first reasoning phase achieves the highest percentage of correctness, indicating that VLMs possess strong perceptual grounding and spatial reasoning ability with \emph{in-frame} visual contents. However, such correctness drops considerably in later phases, reflecting a growing difficulty in maintaining geometric coherence and executing spatial inference \emph{across views}. This trend persists for samples with incorrect final answers, where descriptions remain relatively robust but later reasoning phases collapse. Larger VLMs demonstrate slower performance decay and higher phase stability, implying enhanced integration of cross-view representations. Detailed VLM reasoning paths are shown in Appendix \ref{appendix:E2}.

\subsection{Self-Consistency Prompting}
We perform self-consistency prompting by feeding the multi-view inputs, reasoning path, and final answer back to the same VLM, to verify whether the final answer aligns with VLM's own reasoning. This process assesses VLM's self-coherence and stability on spatial reasoning, revealing contradictions or drift in its reasoning trajectory.

\noindent\textbf{Self-consistency reflects the alignment between reasoning coherence and answer correctness.}
Results in Table~\ref{tab:self_consistency} reveal a strong correspondence between reasoning coherence and final-answer correctness. Across all models, correct answers exhibit substantially higher self-consistency scores, indicating that successful reasoning is typically supported by internally coherent inference chains. In contrast, incorrect answers display much lower self-consistency, suggesting that errors often arise from internal contradictions or unstable relational grounding in reasoning. This discrepancy is particularly pronounced in smaller models, which frequently generate logically inconsistent rationales even when the final answer is correct, reflecting shallow alignment between intermediate reasoning and decision-making. 
These findings show that larger VLMs not only improve task accuracy but also develop more consistent reasoning-to-decision mappings, underscoring self-consistency as a robust indicator of reasoning integrity and cognitive alignment. More details are in Appendix \ref{appendix:E3}.

\section{Implicit VLM Reasoning Analysis}
Implicit analysis complements the explicit analysis, by probing the latent representation of VLMs, such as entropy distribution and representation pattern. While explicit analysis measures what the VLM claims to reason, implicit analysis reveals how its internal representations evolve and where it deteriorates across reasoning phases.

\begin{figure}[ht] \centering \vspace{-0.1in}\includegraphics[width=\linewidth]{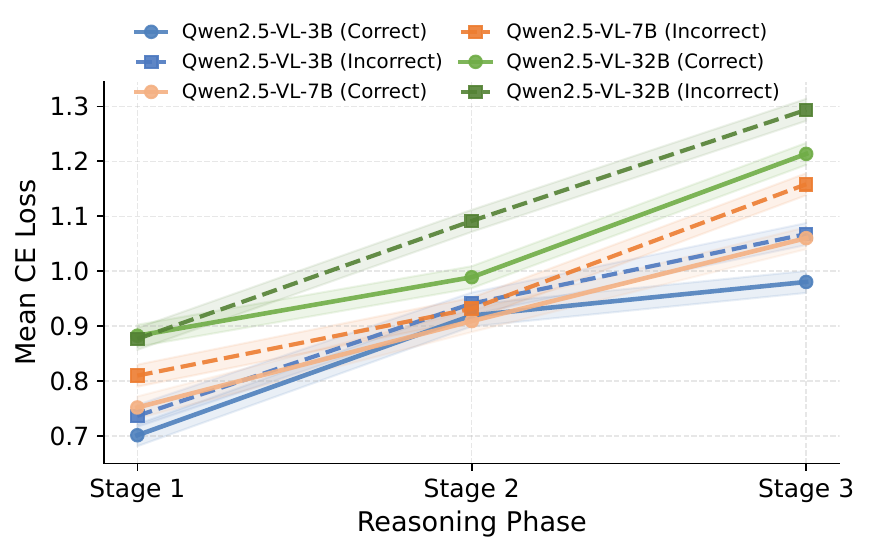} \vspace{-0.3in}\caption{Linear probing analysis on Qwen2.5-VL family} \label{fig:probing} \vspace{-3mm} \end{figure}

\subsection{Linear Probing Analysis}
We conduct linear probing analysis by extracting token logits from the end of each reasoning phase, along with the final answer logits, from all QA samples. A 4-layer MLP probe is then trained using the cross-entropy loss, and analysis is performed by computing the cross-entropy loss between the probe's predicted logits and the VLM's answer logits on the test set. This analysis provides a quantitative measure of how the VLM's latent representations degrade across successive spatial reasoning stages.

\noindent\textbf{VLM representations exhibit progressive degradation of task-relevant information over reasoning phases and model sizes.}
As shown in Figure~\ref{fig:probing}, linear probing results reveal a consistent increase in cross-entropy loss when the reasoning progresses or the model size increases. Early-phase representations exhibit the lowest loss, indicating that perceptual encoding maintains stronger alignment with the final decision signal. As reasoning progresses, the loss rises steadily, suggesting that spatial relations are transformed into higher-level and distributed representations that are less directly aligned with the output space. This effect becomes more pronounced in larger models. Although they achieve higher accuracy, their latent features become increasingly abstract, resulting in higher probe loss. The divergence between correct and incorrect samples also grows in later stages, suggesting that successful reasoning depends on preserving stable information flow from perceptual encoding to decision making, and failed reasoning reflects cumulative degradation of task-relevant representations in reasoning.

\begin{figure}[ht] \centering \includegraphics[width=\linewidth]{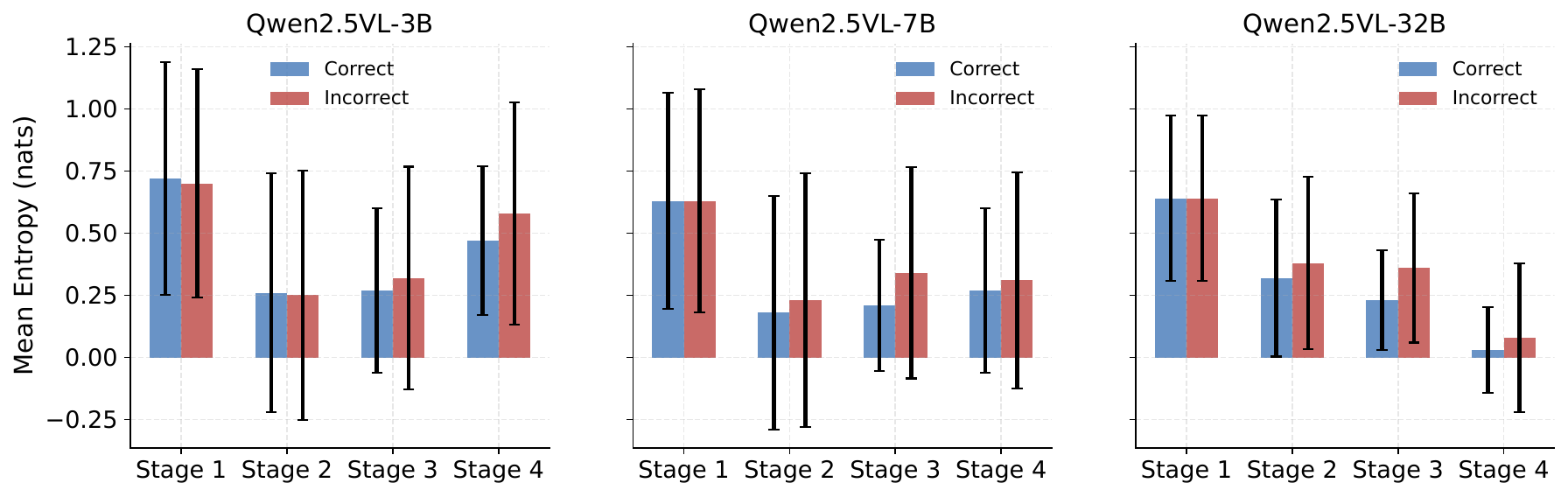} \vspace{-0.2in}\caption{Entropy distribution of Qwen2.5-VL family} \label{fig:entropy} \vspace{-0.2in} \end{figure}

\subsection{Entropy Dynamics Analysis}

We analyze the evolution of entropy across reasoning phases, by computing the mean and variance of token-level logit entropy at each reasoning phase.

\noindent\textbf{Model confidence distinctly separates final answer correctness in reasoning phases.}
As shown in Figure~\ref{fig:entropy}, correct and incorrect reasoning cases exhibit a consistent and well-defined separation in entropy. For model in all sizes, correct reasoning maintains substantially lower entropy, indicating more stable and confident reasoning trajectories. In contrast, incorrect reasoning display persistently higher entropy, reflecting uncertain or conflicting internal representations. 
This separation is most pronounced in larger models, where correct reasoning paths are both low-entropy and tightly distributed, indicating calibrated confidence, while incorrect paths remain diffuse. This finding indicates that entropy patterns strongly modulate spatial reasoning correctness, and that entropy is an effective proxy for diagnosing reliability and correctness in spatial reasoning.

\section{Conclusion of Findings}

Our work presents \textit{ReMindView-Bench}, a cognitively grounded benchmark for evaluating spatial reasoning in multi-view settings. By combining explicit reasoning path analysis with implicit representation probing, we uncover fine-grained mechanisms explaining where and how current VLMs diverge from human-like spatial cognition. Comprehensive analysis shows that VLM limitations primarily stem from insufficient cross-view geometric alignment, unstable inference progression, and weak confidence calibration across reasoning phases.
\begin{itemize}
	\itemsep0em 
	\item Our \textit{explicit analysis} reveal a clear gap between perceptual grounding and geometric inference. VLMs remain accurate in early perceptual encoding but degrade sharply in relational alignment and inferential reasoning. This phase-dependent decay indicates that, although VLMs recognize within-frame relations reliably, they struggle to maintain cross-view coherence. Self-consistency prompting further confirms that stable reasoning chains strongly correlate with correct outcomes. 
	\item Our \textit{implicit analysis} reinforce these findings. Linear probing shows a progressive loss of task-relevant information across reasoning phases, suggesting that spatial signals become increasingly abstract and less accessible. Entropy dynamics reveal persistent confidence separation between correct and incorrect reasoning, indicating systematic uncertainty propagation through reasoning.
\end{itemize}
These findings highlight the need for VLM to maintain coherent spatial representations cross views, calibrate uncertainty dynamically, and integrate explicit and implicit signals to achieve cognitively grounded spatial reasoning.



{
	\small
	\bibliographystyle{ieeenat_fullname}
	\bibliography{main}
}

\appendix

\noindent In this appendix, we provide the following contents:
\begin{itemize}
	\item Details about the VQA design in \textit{ReMindView-Bench} (Appendix~\ref{appendix:B}).
	\item Technical details about \textit{ReMindView-Bench}'s benchmark construction (Appendix~\ref{appendix:C}).
	\item \textit{ReMindView-Bench} evaluation settings (Appendix \ref{appendix:D}).
	\item Details of the explicit analysis of VLM spatial reasoning (Appendix \ref{appendix:E}).
\end{itemize}

\section{VQA Design in \textit{ReMindView-Bench}}
\label{appendix:B}
\subsection{VQA Classification Scheme}
\label{appendix:B1}

ReMindView-Bench systematically evaluates spatial reasoning by decomposing each VQA instance into cognitively grounded dimensions that govern both the visual input and the query formulation. Motivated by core spatial cognition factors introduced in Section 2.2 of the paper's main texts, ReMindView-Bench provides a fine-grained and interpretable space of spatial reasoning tasks. This appendix details the full classification scheme used in ReMindView-Bench.

\noindent\textbf{Vision-side variability}. 
The visual observations in ReMindView-Bench are generated from procedurally synthesized indoor scenes and rendered under diverse geometric and semantic conditions. The variability of the visual input spans four key dimensions, as listed below.

\emph{(1) Room type.}
Five canonical indoor environments are included: dining room, kitchen, bathroom, living room, and bedroom. These room categories differ in spatial layout, functional affordances, and typical object arrangements, introducing natural diversity in occlusion patterns, clutter, and spatial topology.

\emph{(2) Viewpoint spatial pattern.}
Two complementary configurations are used to control the spatial reference frame. The view-centric pattern anchors each observation to the camera’s egocentric perspective, emphasizing perceptual geometry, occlusion variability, and viewpoint-dependent distortions. The object-centric pattern instead anchors each observation to a central scene object, producing allocentric renderings that stabilize the viewpoint and reduce egocentric ambiguity. This dimension determines whether models must rely on egocentric cues or can form viewpoint-invariant spatial representations.

\emph{(3) Level of distance.}
Camera–object distance is discretized into ten fine-grained levels, ranging from very close (level 0) to far-field (level 10). These distance levels modulate spatial proximity, changes in object scale, and the strength of parallax cues. This enables controlled analysis of how models integrate metric information across different spatial regimes.

\emph{(4) Number of visible objects.}
Scenes vary in the total number of visible objects across rendered views. This controls the degree of clutter, relational density, and the working-memory load required to maintain and compare multiple entities. Large numbers of visible objects introduce complex multi-object interactions and increase the chance of distractor interference.

\noindent\textbf{Query-side variability}. 
Each VQA query simultaneously engages several cognitive operations, including relational comparison, reference-frame transformation, and multi-view integration. Every query in ReMindView-Bench is jointly characterized along the following four orthogonal query-side dimensions, each capturing a distinct aspect of spatial reasoning.

\emph{(1) Query type.}
Each question concerns either relative distance, requiring metric comparison between two entities, or relative direction, requiring directional judgments such as left versus right or front versus behind. These two primitives correspond to distinct spatial processing pathways in human cognition.

\emph{(2) Relation type.}
The relational structure underlying each query reflects which entities participate in the comparison. A query may involve a view–object relation between the camera and an object, an object–object relation between two scene objects, or a view–view relation between different viewpoints. This dimension isolates reasoning anchored to the camera, within the scene, or across viewpoints.

\emph{(3) Perspective taking.}
Each query requires either current-view reasoning, in which the answer must be derived from the active camera frame, or shifted reasoning, in which the model must mentally adopt an alternative reference frame, typically tied to an object-centered viewpoint. This dimension evaluates the model’s ability to perform viewpoint transformations analogous to human mental rotation and spatial perspective taking.

\emph{(4) Cross-frame reasoning.}
Queries further differ in whether they can be resolved from a single frame or require integrating information across multiple views. Non-cross-frame queries contain all relevant entities within a single observation. Cross-frame queries distribute relevant entities across multiple views, requiring multi-view fusion, entity correspondence tracking, and global scene integration. This dimension directly measures a model’s ability to consolidate spatial information across views.

Together, these 8 dimensions — 4 on the vision side and 4 on the query side — form a richly cognitively structured classification scheme for evaluating spatial reasoning. By disentangling perceptual variability, relational structure, reference-frame transformations, and multi-view integration, ReMindView-Bench enables detailed diagnosis of model strengths and limitations and provides a principled foundation for future research on cognitively grounded spatial reasoning.

\subsection{More \textit{\textbf{RemindView-Bench}} VQA examples}
\label{appendix:B2}

\begin{figure}[ht] \centering \includegraphics[width=\linewidth]{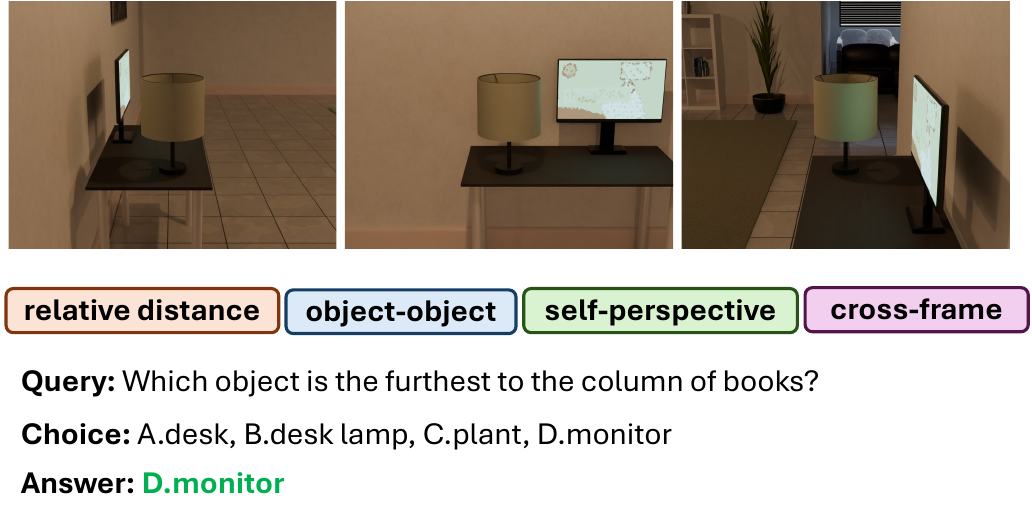} \vspace{-0.3in}\caption{VQA example.} \label{fig:VQA_example_1} \end{figure}

\begin{figure}[ht] \vspace{-0.1in} \centering \includegraphics[width=\linewidth]{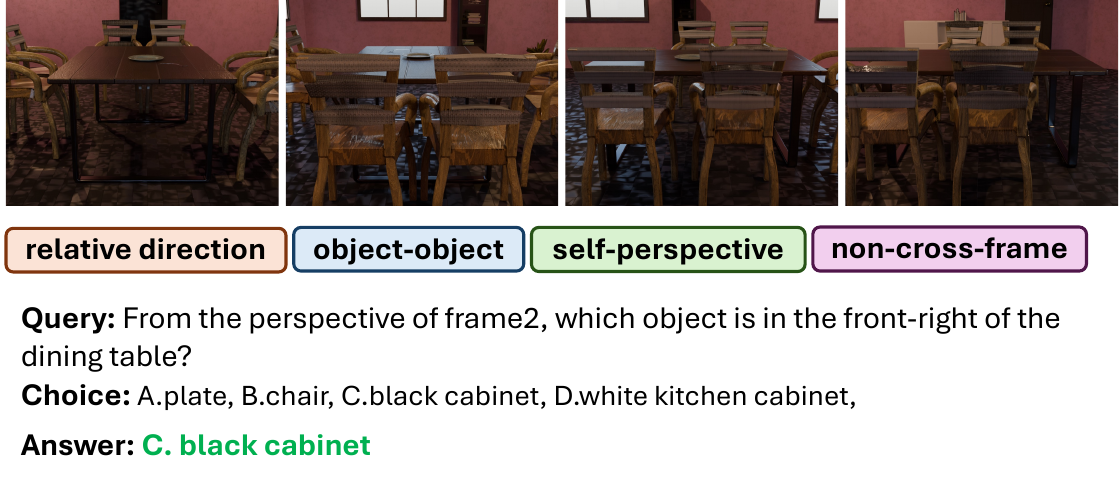} \vspace{-0.3in}\caption{VQA example.} \label{fig:VQA_example_2} \end{figure}

\begin{figure}[ht] \vspace{-0.1in} \centering \includegraphics[width=\linewidth]{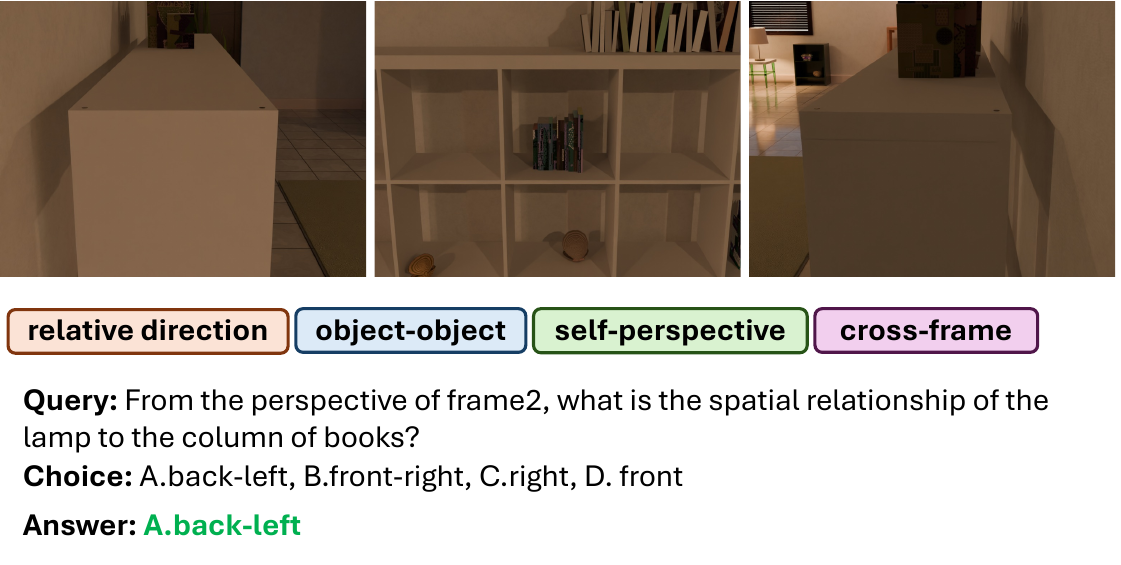} \vspace{-0.3in}\caption{VQA example.} \label{fig:VQA_example_3} \end{figure}

\begin{figure}[ht] \vspace{-0.1in} \centering \includegraphics[width=\linewidth]{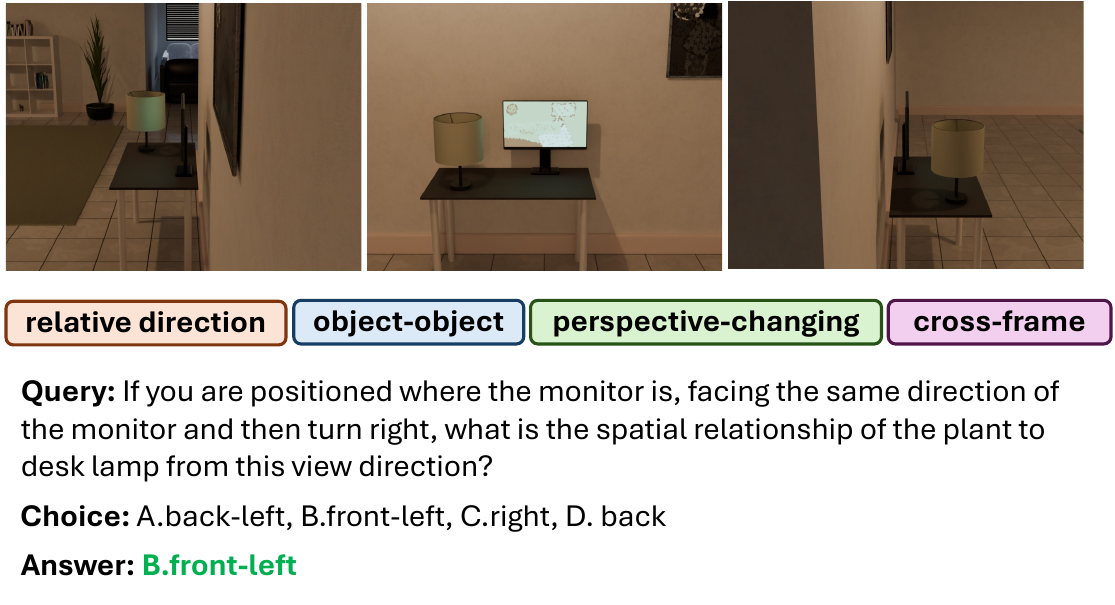} \vspace{-0.3in}\caption{VQA example.} \label{fig:VQA_example_4} \end{figure}

\begin{figure}[ht] \vspace{-0.1in} \centering \includegraphics[width=\linewidth]{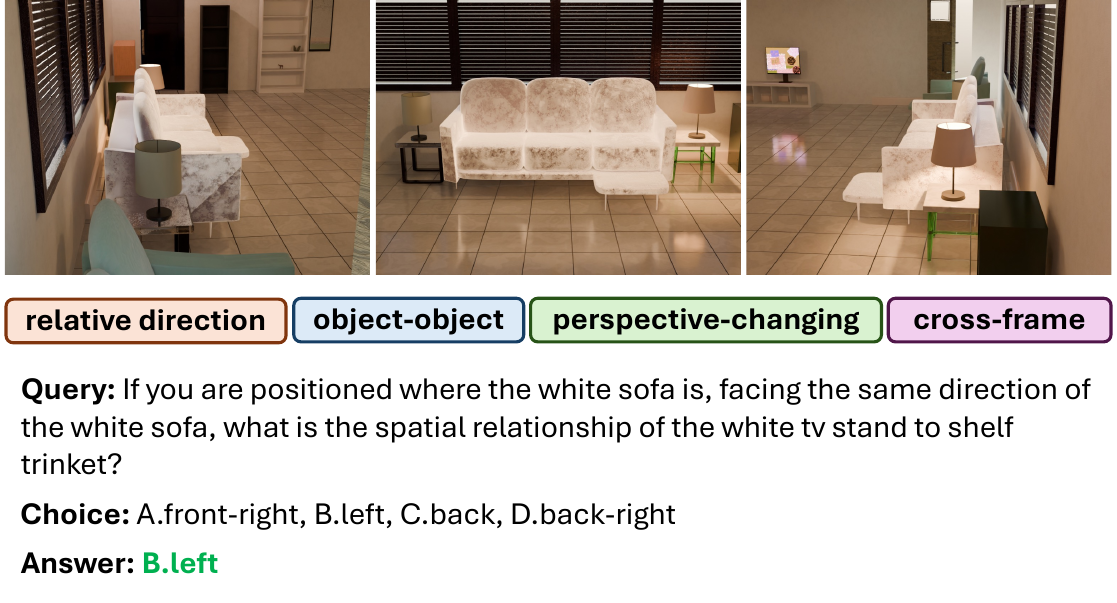} \vspace{-0.3in}\caption{VQA example.} \label{fig:VQA_example_5} \end{figure}

\begin{figure}[ht] \vspace{-0.1in} \centering \includegraphics[width=\linewidth]{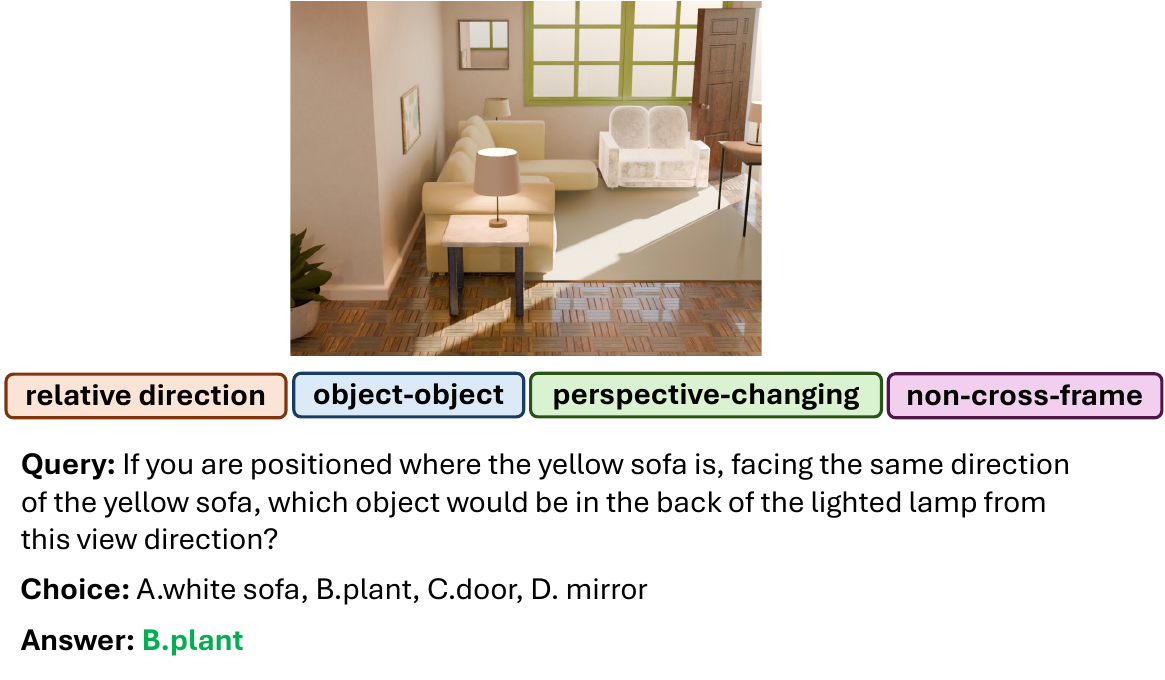} \vspace{-0.3in}\caption{VQA example.} \label{fig:VQA_example_6} \end{figure}

\begin{figure}[ht] \vspace{-0.1in} \centering \includegraphics[width=\linewidth]{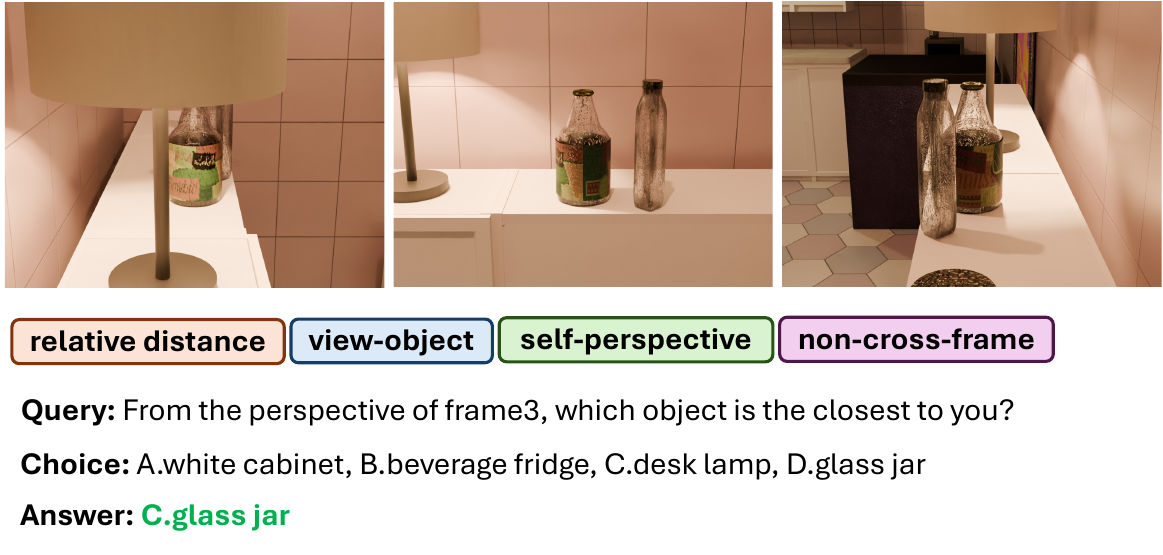} \vspace{-0.3in}\caption{VQA example.} \label{fig:VQA_example_7} \end{figure}

\begin{figure}[ht] \vspace{-0.1in} \centering \includegraphics[width=\linewidth]{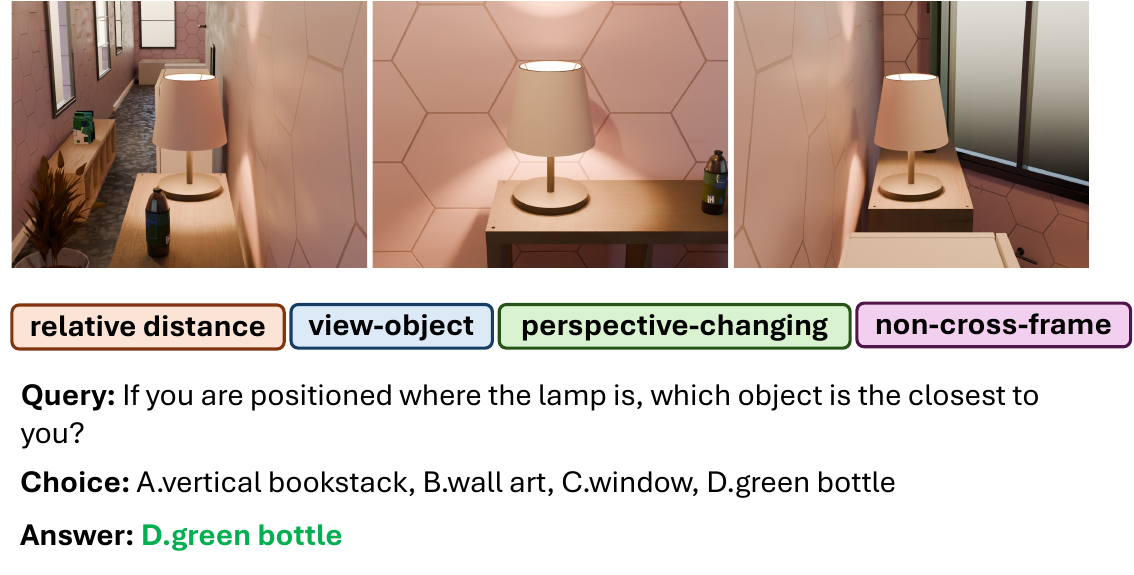} \vspace{-0.3in}\caption{VQA example.} \label{fig:VQA_example_8} \end{figure}

\begin{figure}[ht] \vspace{-0.1in} \centering \includegraphics[width=\linewidth]{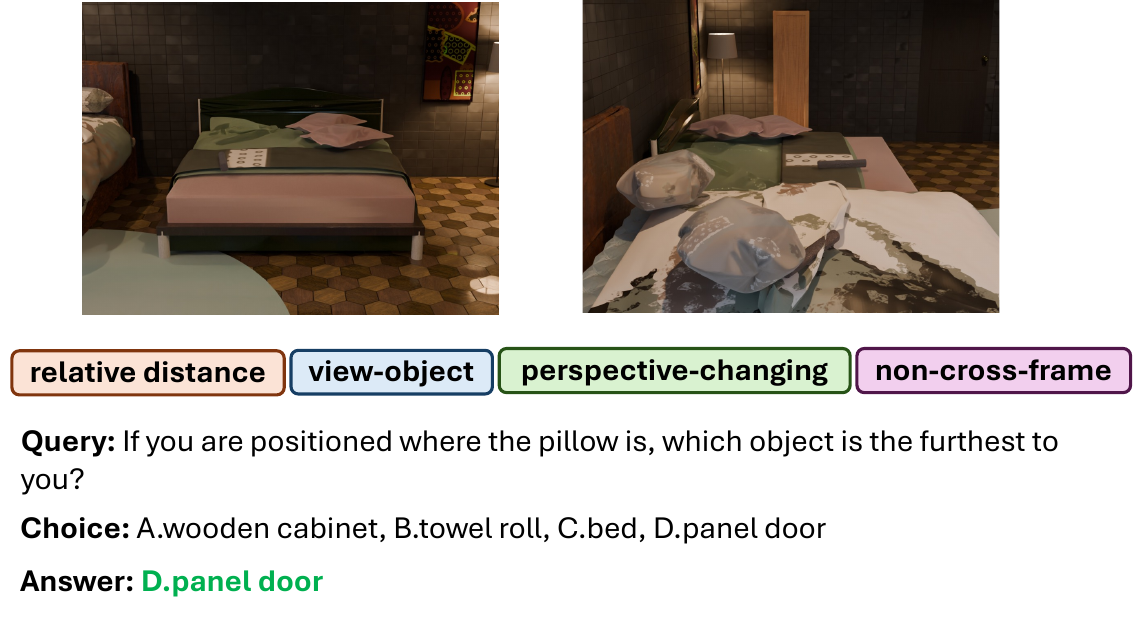} \vspace{-0.3in}\caption{VQA example.} \label{fig:VQA_example_9} \end{figure}

\begin{figure}[ht] \vspace{-0.1in} \centering \includegraphics[width=\linewidth]{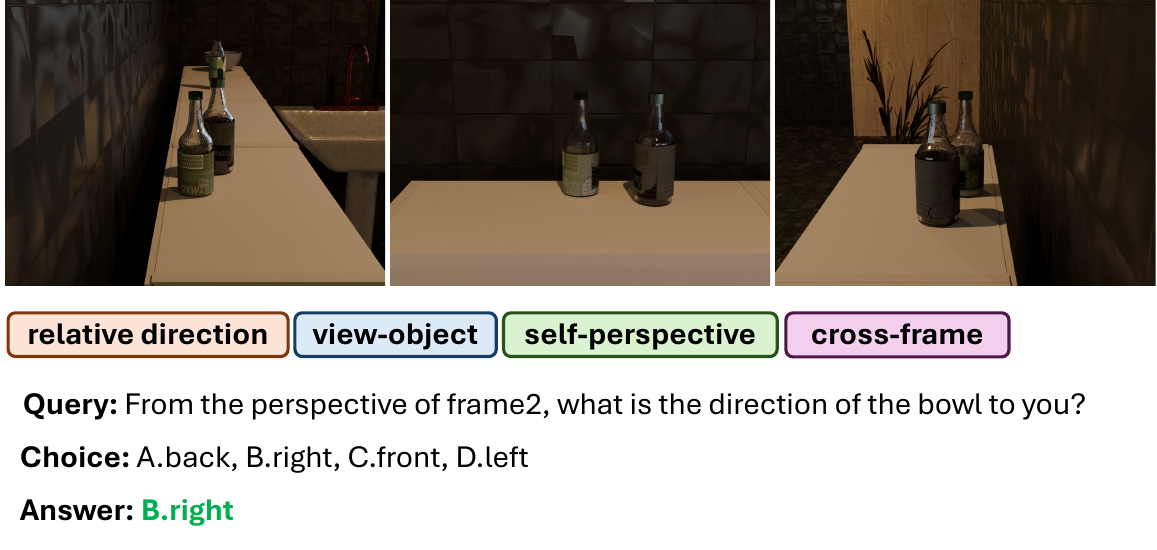} \vspace{-0.3in}\caption{VQA example.} \label{fig:VQA_example_10} \end{figure}

\begin{figure}[ht] \vspace{-0.1in} \centering \includegraphics[width=\linewidth]{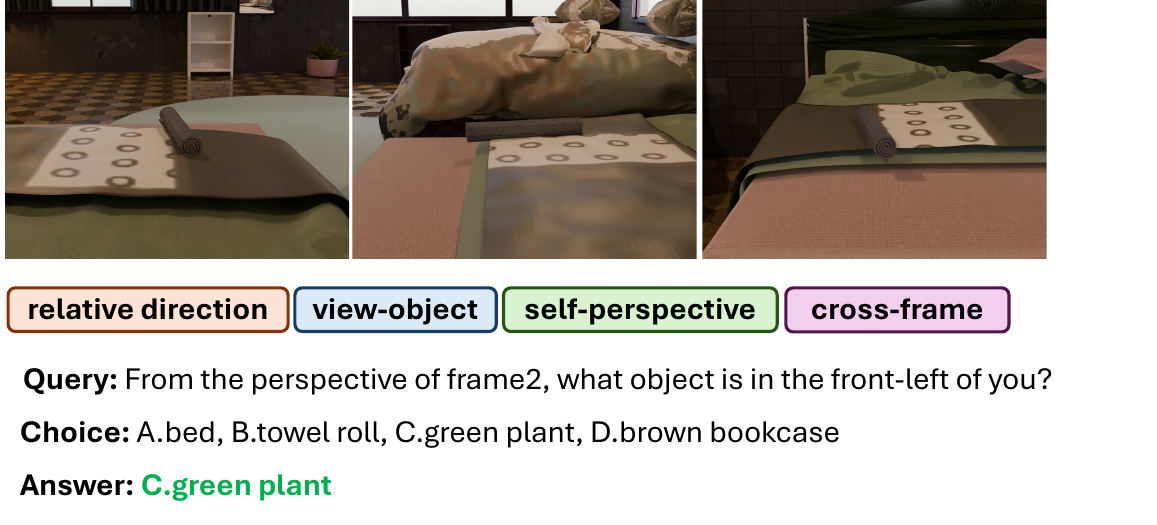} \vspace{-0.3in}\caption{VQA example.} \label{fig:VQA_example_11} \end{figure}

\begin{figure}[ht] \vspace{-0.1in} \centering \includegraphics[width=\linewidth]{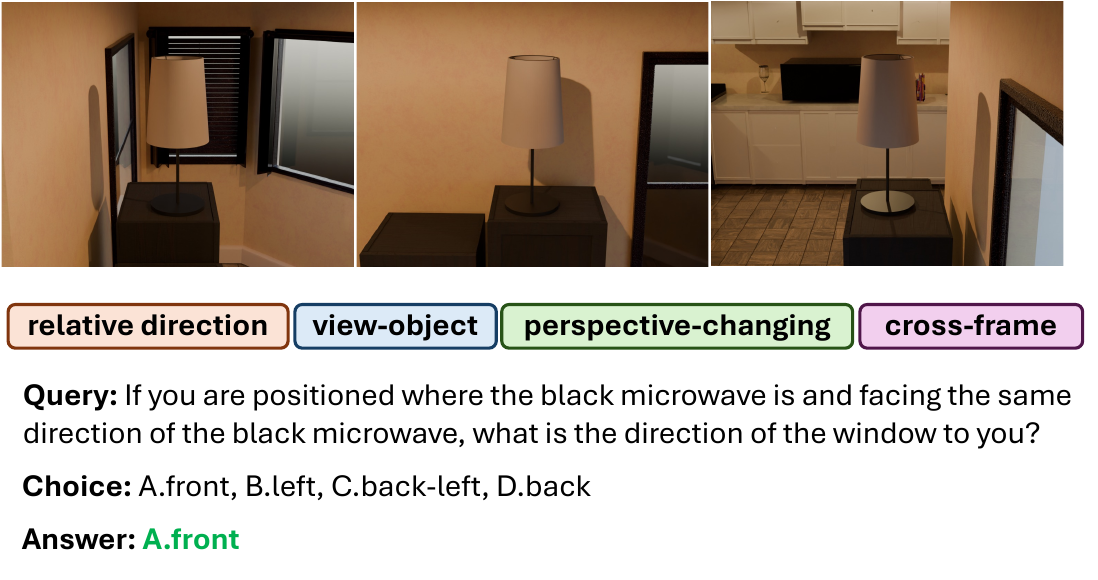} \vspace{-0.3in}\caption{VQA example.} \label{fig:VQA_example_12} \end{figure}

\begin{figure}[ht] \vspace{-0.1in} \centering \includegraphics[width=\linewidth]{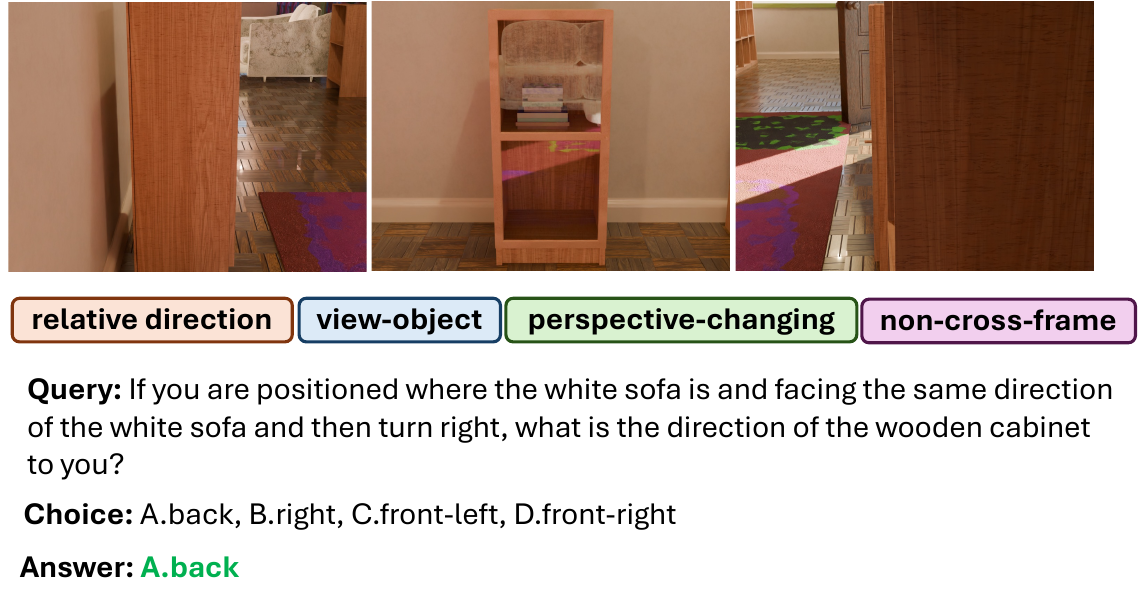} \vspace{-0.3in}\caption{VQA example.} \label{fig:VQA_example_13} \end{figure}

\begin{figure}[ht] \vspace{-0.1in} \centering \includegraphics[width=\linewidth]{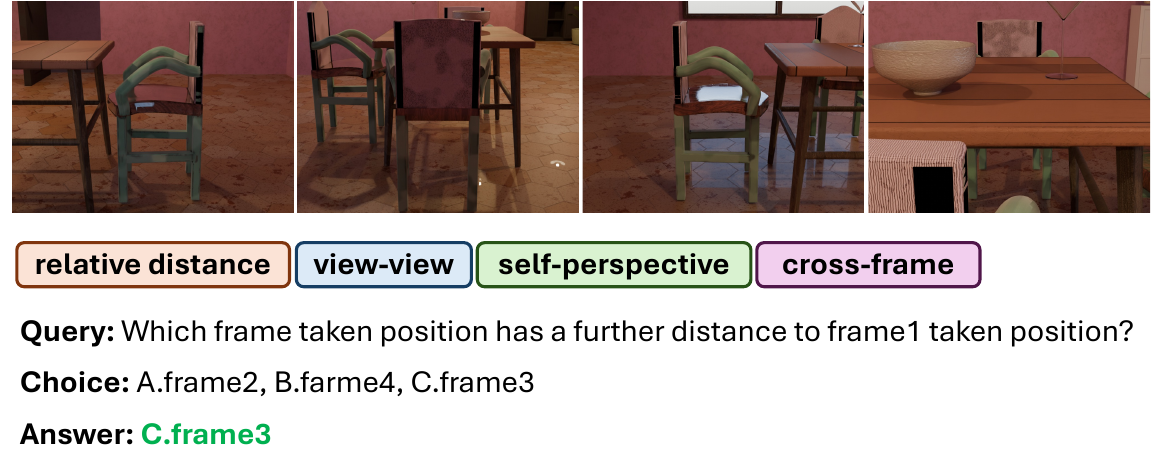} \vspace{-0.3in}\caption{VQA example.} \label{fig:VQA_example_14} \end{figure}

\begin{figure}[ht] \vspace{-0.1in} \centering \includegraphics[width=\linewidth]{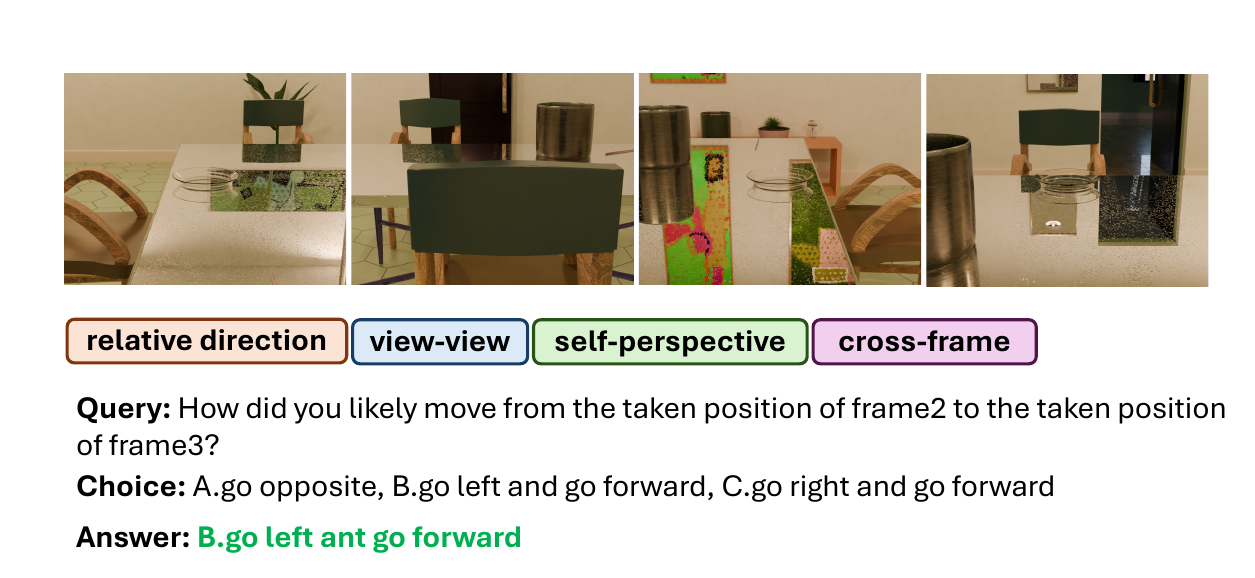} \vspace{-0.3in}\caption{VQA example.} \label{fig:VQA_example_15} \end{figure}

We provide additional VQA examples generated under different prompt templates. Examples are shown in Figures~\ref{fig:VQA_example_1}–\ref{fig:VQA_example_15}. Different combinations of spatial cognitive factors being incorporated are listed in each VQA example.


\subsection{\textbf{\textit{ReMindView-Bench}} statistics}
\label{appendix:B3}

\begin{table}[ht]
	\centering
	\begin{tabular}{l@{\hskip 1em}r}
		\toprule
		\textbf{Statistic} & \textbf{Count (\%)} \\
		\midrule
		\multicolumn{2}{l}{\textit{Visual Data}} \\ 
		\quad Unique sparse room number & 50 \\
		\quad Unique dense room number & 50 \\
		\quad Unique object number & 444 \\ 
		\quad Unique multi-view image set number & 6{,}977 \\ 
		\midrule
		\multicolumn{2}{l}{\textit{Cross-Frame Reasoning}} \\
		\quad Cross-frame VQA number & 40{,}644 (81.3\%) \\
		\quad Non cross-frame VQA number & 9{,}356 (18.7\%) \\
		\midrule
		\multicolumn{2}{l}{\textit{Perspective Taking}} \\
		\quad Perspective-changing VQA number & 21{,}250 (53.1\%) \\
		\quad Self-perspective VQA number & 18{,}750 (46.9\%) \\
		\midrule
		\multicolumn{2}{l}{\textit{Number of Choices}} \\
		\quad 2 choices & 9{,}258 (18.5\%) \\
		\quad 3 choices & 13{,}868 (27.7\%) \\
		\quad 4 choices & 26{,}874 (53.8\%) \\
		\bottomrule
	\end{tabular}
	\vspace{-0.1in}
	\caption{Benchmark statistics for visual data, cross-frame reasoning, perspective taking and number of choices}
	\label{tab:dataset_statistic_1}
\end{table}

\begin{table}[ht]
	\centering
	\begin{tabular}{llr}
		\toprule
		\textbf{Query Type} & \textbf{Relation Type} & \textbf{Count (\%)} \\
		\midrule
		\multirow{3}{*}{Relative Direction}
		& object--object & 14{,}624 (29.2\%) \\
		& view--object   & 12{,}000 (24.0\%) \\
		& view--view     & 5{,}000 (10.0\%) \\
		\midrule
		\multirow{3}{*}{Relative Distance}
		& object--object & 5{,}376 (10.8\%) \\
		& view--object   & 8{,}000 (16.0\%) \\
		& view--view     & 5{,}000 (10.0\%) \\
		\midrule
		\multicolumn{2}{l}{\textbf{Total}} & \textbf{50{,}000 (100\%)} \\
		\bottomrule
	\end{tabular}
	\vspace{-0.1in}
	\caption{Statistics of benchmark query types, grouped by relation type}
	\label{tab:dataset_statistic_2}
\end{table}

\begin{figure}[ht] \vspace{-0.1in} \centering \includegraphics[width=0.9\linewidth]{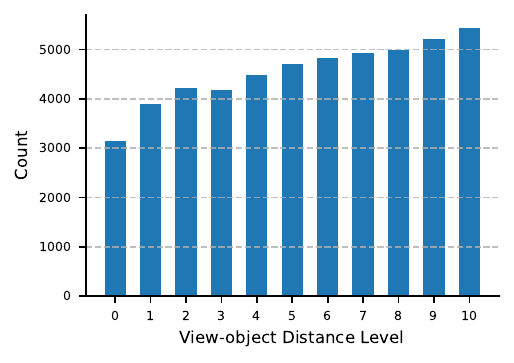} \vspace{-0.1in}\caption{Distribution of the benchmark's object-view distance levels} \label{fig:distance_level_distribution} \vspace{-0.1in} \end{figure}

\begin{figure}[ht]  \centering \includegraphics[width=\linewidth]{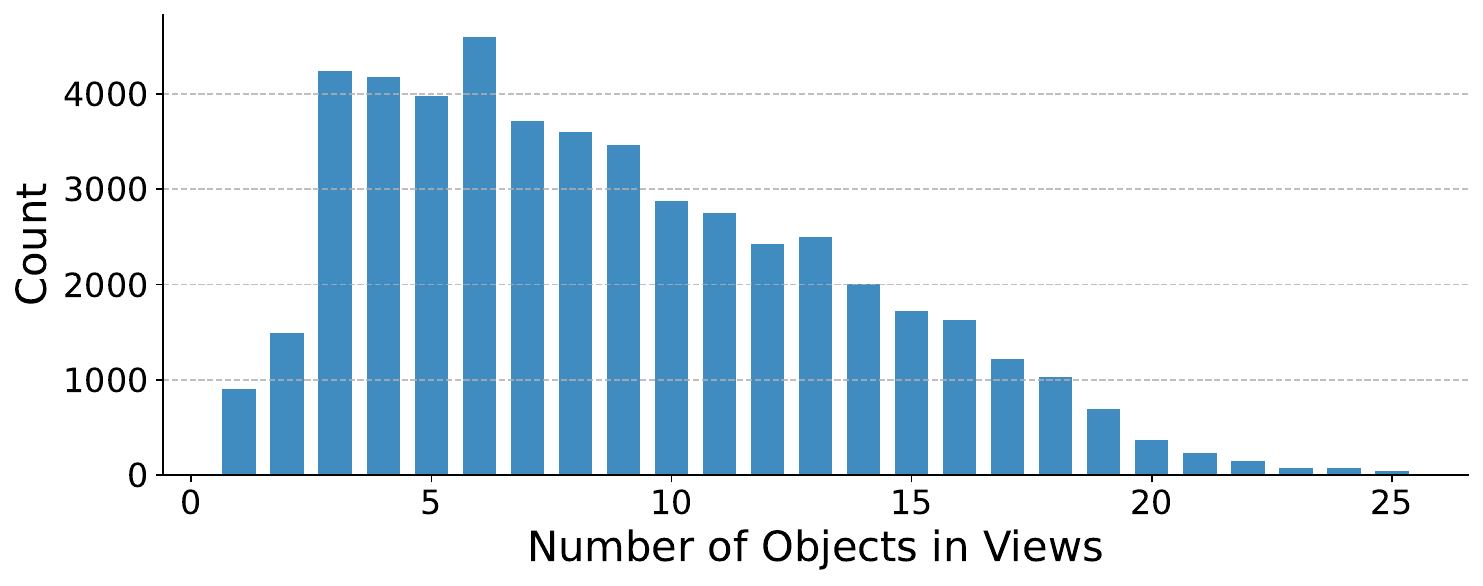} \vspace{-0.3in}\caption{Distribution of the benchmark's number of objects in views} \label{fig:object_number_distribution} \vspace{-0.1in} \end{figure}

Table~\ref{tab:dataset_statistic_1} provides an overview of the benchmark’s core visual and reasoning statistics. The dataset contains 100 unique room instances and 444 distinct objects, forming 6,977 multi-view image sets that span highly varied indoor layouts. In terms of reasoning complexity, 81.3\% of all VQA queries require cross-frame integration, indicating that most tasks cannot be solved from a single viewpoint and instead demand multi-view relational consistency. Perspective-taking difficulty is also balanced, 53.1\% of questions involve perspective-changing reasoning, while 46.9\% remain within an egocentric reference frame. The distribution of answer-choice formats ranges from 2 to 4 options, with the majority adopting 4-choice settings, which reduces chance-level correctness and contributes to higher task difficulty. Table~\ref{tab:dataset_statistic_2} further details the relational breakdown of the 50,000 queries. Both relative-direction and relative-distance tasks appear across object–object, view–object, and view–view configurations, ensuring that the benchmark emphasizes a variety of relation type in queries.

Figures~\ref{fig:distance_level_distribution} and~\ref{fig:object_number_distribution} illustrate the geometric and structural variability of the visual data. The distribution of view–object distance levels spans 0 to 10 discrete bins, roughly covering the close to far range. Most distance bins exhibit comparable sample sizes, with a mild increase toward farther distances, ensuring diverse depth cues and parallax patterns across scenes. The number of objects visible in each view follows a long-tailed distribution, most images contain between 3 and 9 objects, while dense scenes with over 20 objects also appear. This range of object densities introduces natural clutter, frequent occlusion, and complex multi-object interactions that further challenge spatial reasoning models. Together, these statistics demonstrate that \textit{ReMindView-Bench} not only provides large-scale multi-view data but also systematically incorporates geometric diversity, relational heterogeneity, and multi-object scene complexity essential for evaluating fine-grained 3D spatial reasoning.

\section{Technical Details about \textit{ReMindView-Bench}'s Benchmark Construction}
\label{appendix:C}
\subsection{Settings of Scene Generation Constraints in Infinigen}
\label{appendix:C1}

We generate the 3D indoor scenes in \textit{ReMindView-Bench} using the Infinigen procedural generation framework \cite{raistrick2024infinigen}, which provides a fully parametric and constraint-driven system for synthesizing diverse yet physically and semantically coherent scenes. The generator combines procedural asset programs with a high-level constraint specification language and a simulated-annealing solver to satisfy geometric, ergonomic, and functional relationships in indoor spaces. This enables controlled variation across room layout, object composition, and clutter level while preserving physical plausibility.

\noindent\textbf{Scene type and architectural structure.} For each benchmark scene, we begin by sampling 10 in each five room types (living room, dining room, kitchen, bedroom, and bathroom) in spares and dense settings, and generate its architectural shell—walls, floor, ceiling, doors, and windows—using Infinigen’s default procedural floor-plan module. The floor-plan solver optimizes a weighted objective combining room area, aspect ratio, convexity, free-space constraints, and wall conciseness. These objectives ensure the resulting rooms maintain realistic geometric proportions and functional accessibility suitable for indoor spatial reasoning tasks.

\noindent\textbf{Object density and semantic placement.} To modulate scene clutter and relational complexity, we vary Infinigen’s object-density parameters from low (0-0.3) to high (0.7-1.0), corresponding to sparse and dense scene configurations. The arrangement process follows Infinigen’s hierarchical solving strategy. Large functional furniture (e.g., beds, sofas, tables) are first placed under ergonomic and accessibility constraints, including minimum walkable free-space, symmetry around central items, and semantic adjacency relations (e.g., chairs positioned and facing around a dining table). Once the macro layout is stabilized, medium and small objects are introduced under support-surface, attachment, and spatial-semantic constraints (e.g., lamps must rest on horizontal surfaces; small items must not intersect or float). To further differentiate sparse and dense scenes, we adjust the likelihood that medium and small objects are instantiated on, against, or inside large objects. Sparse scenes decrease the probability (0.2), whereas dense scenes increase their probability (0.8). This staged optimization ensures controllable clutter and rich relational structure while preventing physically implausible overlaps.

\noindent\textbf{Room-specific semantic constraints.} Different room types trigger different constraint bundles. For example, dining rooms activate symmetry and spacing constraints for tables and seating; kitchens require collocation of appliances and free-space in front of doors; bathrooms enforce clearance around sinks and toilets; living rooms prioritize seating clusters and visibility relations. These room-type–conditioned constraints produce semantically coherent scenes aligned with typical human expectations while maintaining procedural variability.

\noindent\textbf{Constraint-based solver configuration.} We adopt Infinigen’s simulated-annealing–based constraint solver to enforce global plausibility while allowing stochastic diversity. During optimization, the solver alternates discrete moves (adding, deleting, or resampling object parameters) and continuous moves (translation/rotation along the object’s allowable degrees of freedom). To avoid excessive solver time while maintaining high-quality layouts, we restrict the maximum number of solver steps to 1,000 and apply the default cooling schedule from 0.25 to 0.001. We further enable BVH caching, evaluation caching, and move filtering—features shown to significantly accelerate convergence without sacrificing arrangement quality.

\noindent\textbf{Integration with Blender rendering.} After optimization, the finalized scene mesh is imported into Blender, where we place cameras and compute object metadata for later VQA generation. Because Infinigen ensures non-intersecting, physically plausible arrangements, the resulting rendered views exhibit natural occlusions, diverse depth ordering, and clutter patterns essential for evaluating multi-view spatial reasoning.

\subsection{Viewpoint Settings for View Rendering}
\label{appendix:C2}

\begin{figure}[ht] \centering \includegraphics[width=\linewidth]{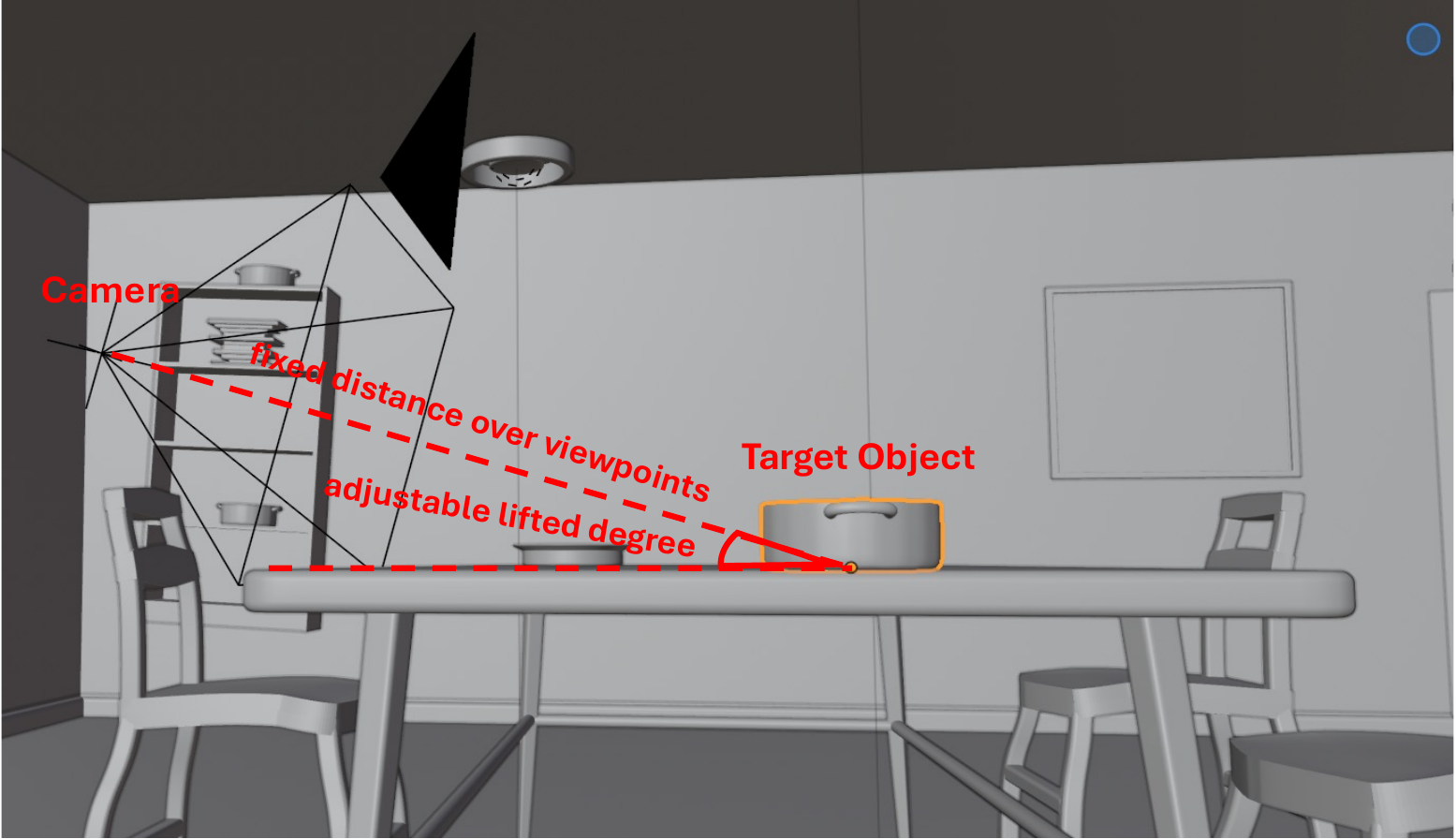} \vspace{-0.25in}\caption{Illustration of general viewpoint settings in Blender} \label{fig:view_angle} \vspace{-0.1in} \end{figure}

\textbf{General settings.} To produce structured and geometrically controlled observations for multi-view spatial reasoning, we adopt a principled camera-placement strategy in Blender. For each target object, the rendering process begins by positioning the cameras in  view centric or object centric spatial relationship with a fixed radial distance from the object while allowing an adjustable lifted angle in a range (25-35 degree). This configuration ensures that views every VQA frame set maintains the same object–view distance. The elevation of the camera is kept relatively stable across renders to approximate human-eye height and to avoid distortions caused by large vertical offsets. As shown in Figure~\ref{fig:view_angle}, this setup yields viewpoint variations that isolate geometric reasoning from unconstrained camera motion.

\noindent\textbf{The design of object-view distance levels.} To provide multi-scale spatial cues aligned with principles from human spatial cognition, we discretize the camera-object distance into concentric levels that span near-field to far-field viewpoints. Cognitive studies show that humans rely on distinct spatial cues across viewing distances: near distances accentuate fine-grained metric relations and occlusion patterns, while farther viewpoints support allocentric alignment and schema-based layout reasoning. To adapt distance to the scale of each scene in this perspective, we define the canonical object-view distance level using the object’s XY-plane diagonal $D$ and a scale factor $\alpha = \sqrt{\text{room size} / \text{object size}}$, which reflects the cognitive intuition that perceptual discriminability depends on the ratio between object and scene size. Parameterizing distance using a discrete percentage variable $l$ that takes values from $0$ to $1$ in increments of $0.1$ (corresponding to distance levels 0 through 10), and applying an offset of $1.5$ to ensure a minimum valid object-view separation, we obtain the boundary function $d(n) = \big[n(\alpha - 1.5) + 1.5\big] D$. 

\begin{figure}[ht] \centering \includegraphics[width=\linewidth]{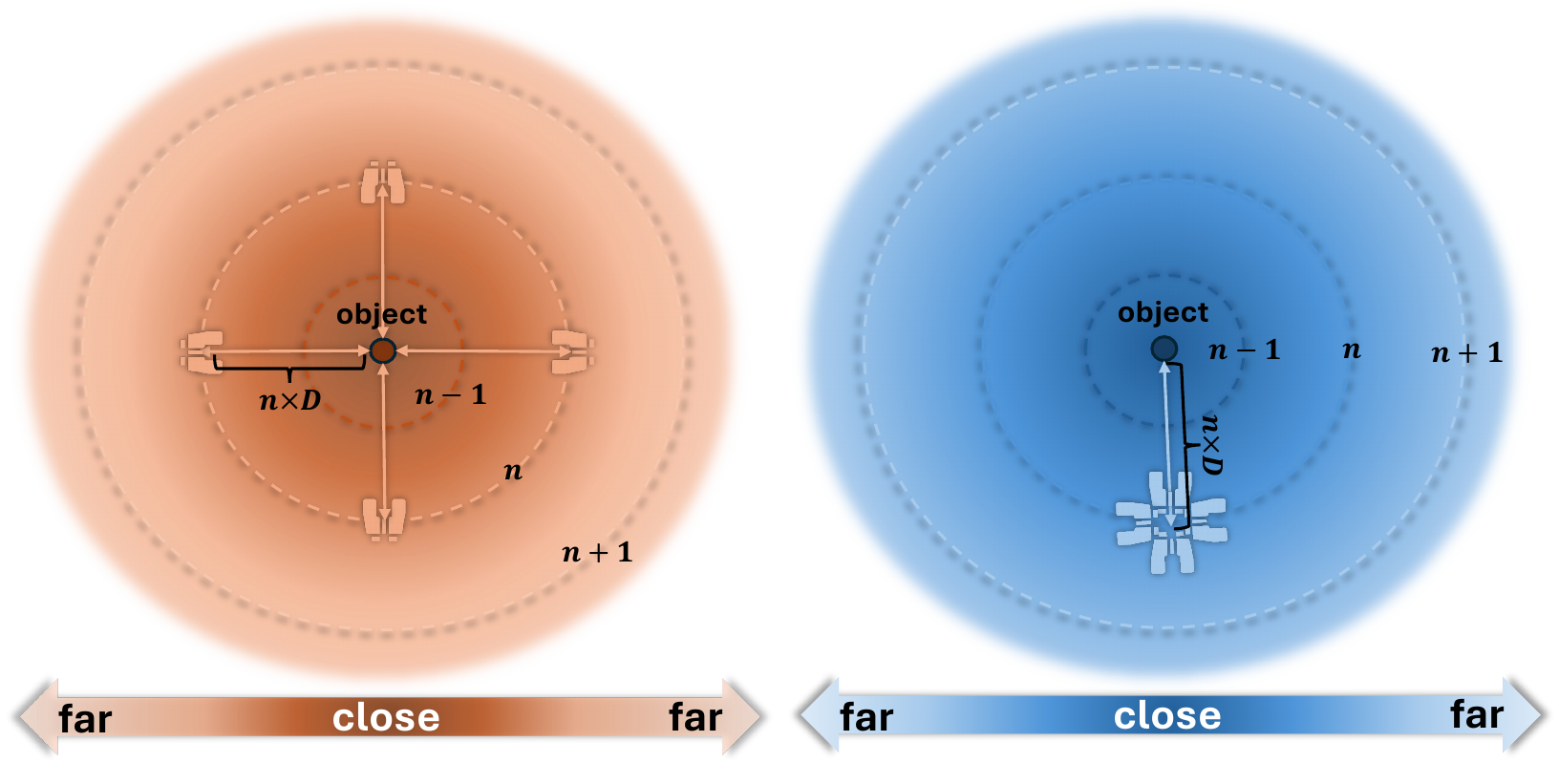} \vspace{-0.2in}\caption{Illustration of object-view distance levels. The right figure shows settings for object-centric spatial patterns, and the left figure shows settings for view-centric spatial patterns.} \label{fig:viewpoint_setting}  \end{figure}

As shown in Figure \ref{fig:viewpoint_setting}, this formulation produces a cognitively grounded set of multi-scale viewpoints that probe how VLMs integrate spatial information across systematic variations in perceptual distance.

\begin{figure}[ht] \vspace{-0.1in} \centering \includegraphics[width=\linewidth]{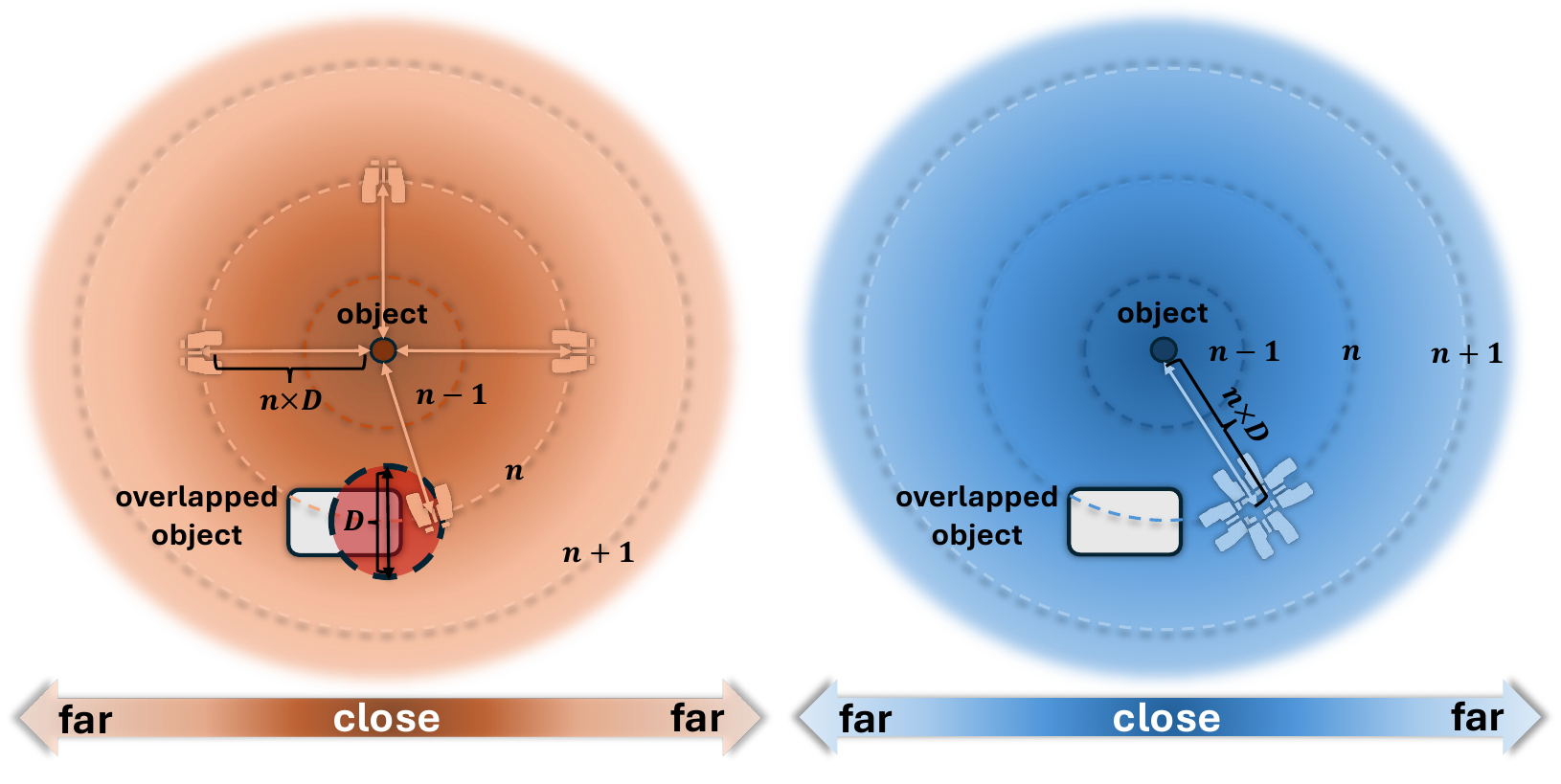} \vspace{-0.2in}\caption{Viewpoint searching strategy illustration} \label{fig:viewpoint_setting_overlap} \vspace{-0.1in} \end{figure}

\begin{figure}[ht] \centering \includegraphics[width=\linewidth]{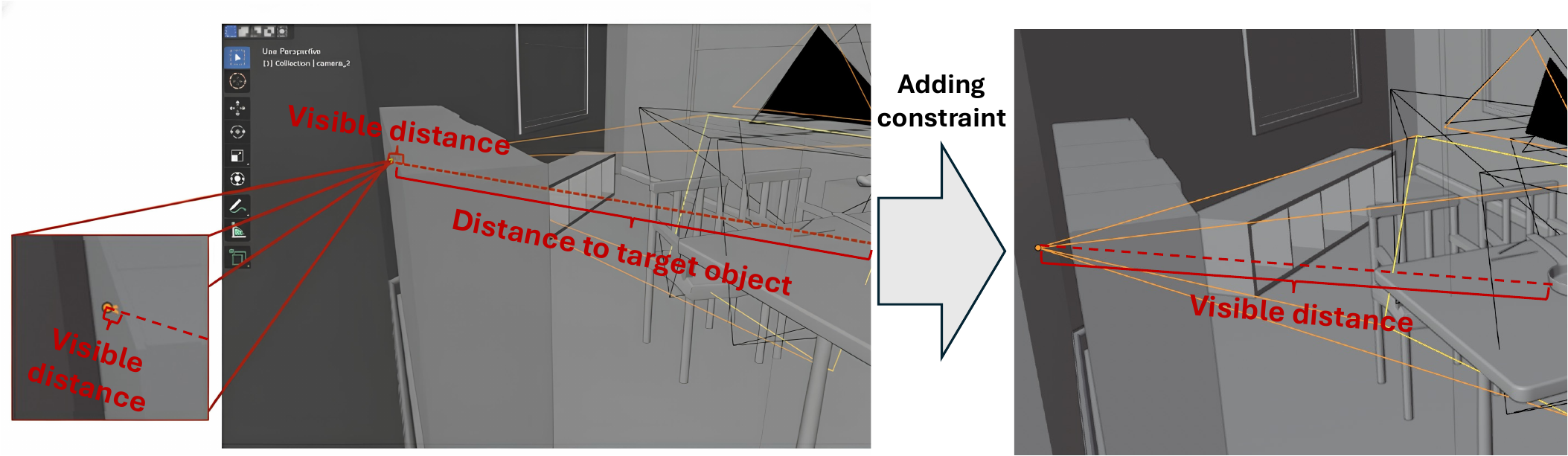} \vspace{-0.15in}\caption{Illustration of viewpoint blocking constraints in Blender} \label{fig:view_block} \vspace{-0.1in} \end{figure}

\noindent\textbf{Viewpoint searching strategy for occluded or overlapping cases.} To ensure that each rendered viewpoint remains spatially valid and visually informative, we employ a constrained viewpoint-searching strategy centered on the target object, as illustrated in Figure~\ref{fig:viewpoint_setting_overlap}. Given a predefined distance level $nD$, the camera is first placed at the corresponding radial position. If the line of sight is blocked by an occluding or overlapping object, the search procedure differs by spatial pattern: in the object-centric setting, we search within a circular region of diameter $D$ around the initial viewpoint; in the view-centric setting, we iteratively scan angular offsets along the same radial trajectory. This strategy preserves geometric consistency with the distance-level formulation while selecting the nearest unobstructed viewpoint. As shown in Figure~\ref{fig:view_block}, it enables the generation of clean and consistent multi-view supervision critical for evaluating spatial reasoning.

\subsection{Details of VQA Generation}
\label{appendix:C3}

\begin{figure}[ht] \centering \includegraphics[width=\linewidth]{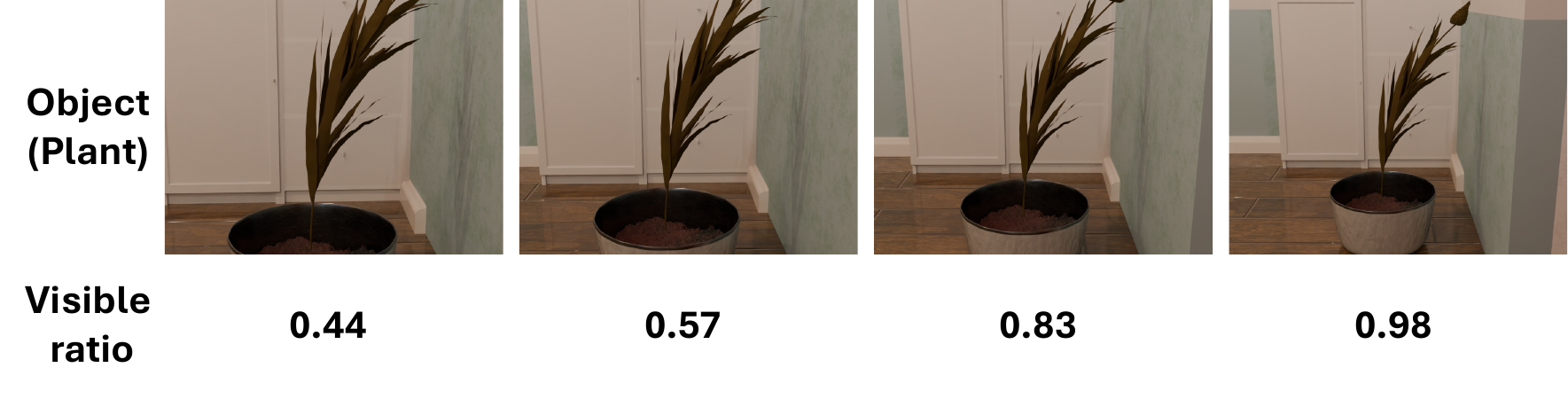} \vspace{-0.2in}\caption{Object visibility ratio example} \label{fig:visible_ratio} \end{figure}

We generate spatial VQA queries by instantiating a set of 22 templates using geometric metadata extracted from Blender. For each rendered frame set, we obtain the set of visible objects along with the camera and object positions and poses, which together provide the geometric information required to compute relative directions and distances and to derive both the answer choices and the ground-truth label. The visible object list is determined through a volumetric occlusion–based visibility test. Each object’s 3D mesh is uniformly sampled at 1000 points, and for every sampled point we cast rays from the camera’s perception field using Blender’s ray-tracing engine. A point is considered visible if it's on the ray tracing fields. The proportion of visible points defines the object’s non-occluded visibility ratio, and objects whose ratio exceeds a predefined threshold (0.2) are retained as visible as example of visible ratio of a plant shown in Figure~\ref{fig:visible_ratio}. This volumetric approach provides a robust estimate of true visibility in cluttered scenes, avoiding false positives caused by small or tangentially exposed surface regions.

\noindent\textbf{Generation of multi-choice questions (MCQs).} Based on the extracted metadata, we instantiate each query using the predefined templates shown in Table~\ref{tab:question_templates}. Object-related placeholders are populated using the list of visible objects in each frame, while direction-related placeholders are sampled from eight canonical horizontal directions including `front', `behind', `left', `right', `front-left', `front-right', `back-left', and `back-right'. To avoid ambiguous, objects exhibiting large vertical overlap-computed from their 3D bounding boxes and spatial positions—are mutually excluded during the MCQ generation process.

\begin{table*}[t]
	\centering
	\small
	\renewcommand{\arraystretch}{1.3} 
	\setlength{\tabcolsep}{6pt}
	
	\renewcommand{\tabularxcolumn}[1]{>{\raggedright\arraybackslash}m{#1}}
	
	\begin{tabularx}{\textwidth}{
			>{\centering\arraybackslash}m{0.12\textwidth}  
			>{\centering\arraybackslash}m{0.16\textwidth}  
			X                                               
		}
		\toprule
		\textbf{Relationship} & \textbf{Query Type \newline (Perspective-taking)} & \multicolumn{1}{c}{\textbf{Example Question Templates}} \\
		\midrule
		
		\multirow{4}{*}{\textbf{View--View}} 
		& \textbf{Relative Distance}
		& 1. Which frame's taken position is closer to the taken position of frame\textless{}frame\_num\_1\textgreater{}? \newline
		2. Which frame's taken position is further from the taken position of frame\textless{}frame\_num\_1\textgreater{}? \\
		\cmidrule{2-3}
		
		& \textbf{Relative Direction}
		& 1. How did you likely move from the taken position of frame\textless{}frame\_num\_1\textgreater{} to that of frame\textless{}frame\_num\_2\textgreater{}? \newline
		2. Which frame is taken at the position \textless{}direction\textgreater{} of the taken position of frame\textless{}frame\_num\_1\textgreater{}? \\
		\midrule
		
		\multirow{13}{*}{\textbf{View--Object}} 
		& \textbf{Relative Distance \newline (SP)}
		& 1. From the perspective of frame\textless{}frame\_num\_1\textgreater{}, which object is the closest to you? \newline
		2. From the perspective of frame\textless{}frame\_num\_1\textgreater{}, which object is the furthest from you? \\
		\cmidrule{2-3}
		
		& \textbf{Relative Distance \newline (PC)}
		& 1. If you are positioned where \textless{}object\_1\textgreater{} is, which object is the closest to you? \newline
		2. If you are positioned where \textless{}object\_1\textgreater{} is, which object is the furthest from you? \\
		\cmidrule{2-3}
		
		& \textbf{Relative Direction \newline (SP)}
		& 1. From the perspective of frame\textless{}frame\_num\_1\textgreater{}, what is the direction of \textless{}object\_1\textgreater{} relative to you? \newline
		2. From the perspective of frame\textless{}frame\_num\_1\textgreater{}, which object is in the \textless{}direction\textgreater{} of you? \\
		\cmidrule{2-3}
		
		& \textbf{Relative Direction \newline (PC)}
		& 1. If you are positioned where \textless{}object\_1\textgreater{} is and facing the same direction, what is the direction of \textless{}object\_2\textgreater{} to you? \newline
		2. If you are positioned where \textless{}object\_1\textgreater{} is and facing the same direction, which object is in the \textless{}direction\textgreater{} of you? \newline
		3. If you are positioned where \textless{}object\_1\textgreater{} is, facing the same direction, and then turn \textless{}direction\textgreater{}, what is the direction of \textless{}object\_2\textgreater{} to you? \newline
		4. If you turn \textless{}direction\_1\textgreater{} from the view of \textless{}object\_1\textgreater{}, which object is in the \textless{}direction\_2\textgreater{} of you? \\
		\midrule
		
		\multirow{11}{*}{\textbf{Object--Object}}
		& \textbf{Relative Distance \newline (SP)}
		& 1. Which object is the closest to \textless{}object\_1\textgreater{}? \newline
		2. Which object is the furthest from \textless{}object\_1\textgreater{}? \\
		\cmidrule{2-3}
		
		& \textbf{Relative Direction \newline (SP)}
		& 1. From the perspective of frame\textless{}frame\_num\_1\textgreater{}, which object is in the \textless{}direction\textgreater{} of \textless{}object\_1\textgreater{}? \newline
		2. From the same perspective, what is the spatial relationship of \textless{}object\_2\textgreater{} to \textless{}object\_1\textgreater{}? \\
		\cmidrule{2-3}
		
		& \textbf{Relative Direction \newline (PC)}
		& 1. If positioned at \textless{}object\_1\textgreater{}, facing its direction, which object is in the \textless{}direction\textgreater{} of \textless{}object\_2\textgreater{}? \newline
		2. If positioned at \textless{}object\_1\textgreater{}, what is the spatial relationship of \textless{}object\_2\textgreater{} to \textless{}object\_3\textgreater{}? \newline
		3. If positioned at \textless{}object\_1\textgreater{} and turn \textless{}direction\_1\textgreater{}, which object is in the \textless{}direction\_2\textgreater{} of \textless{}object\_2\textgreater{}? \newline
		4. After turning \textless{}direction\textgreater{} from \textless{}object\_1\textgreater{}, what is the spatial relationship of \textless{}object\_2\textgreater{} to \textless{}object\_3\textgreater{}? \\
		\bottomrule
	\end{tabularx}
	
	\caption{Question templates categorized by the type of spatial relationship, and query type with perspective-taking (PC for perspective-changing and SP for self-perspective) conditions.}
	\label{tab:question_templates}
\end{table*}

\noindent\textbf{Ground-truth derivation.} The ground-truth answer for each instantiated query is computed directly from the metadata. Directional relations are obtained from the orientation and position of the cameras or objects, while distance relations are determined from Euclidean distances between object centers or camera locations. To construct the multiple-choice format, we begin with the set of visible objects and retain only those whose spatial relations are unambiguous with respect to the template. Specifically, the objects with overlapping direction description (`right' and `front-right') and the object with same or similar label (`brown sofa' and `black sofa') are mutual excluded in the generation process.

\noindent\textbf{Quality control.} A final quality-control stage filters out queries that fail grounding or consistency checks. We discard questions with ambiguous direction labels, distance ties, view-dependent contradictions, or template conditions that fail under the specific scene geometry. This end-to-end process ensures that the benchmark contains only well-structured, cognitively meaningful spatial reasoning queries that accurately reflect the underlying 3D environment.

\section{\textit{ReMindView-Bench} Evaluation Settings}
\label{appendix:D}

\begin{table*}[t]
	\centering
	\scriptsize 
	\renewcommand{\arraystretch}{1.3}
	\setlength{\tabcolsep}{8pt} 
	
	\renewcommand{\tabularxcolumn}[1]{m{#1}}
	
	\begin{tabularx}{\textwidth}{
			>{\centering\arraybackslash}m{0.15\textwidth} 
			>{\raggedright\arraybackslash}X                 
		}
		\toprule
		\textbf{Pattern} & \multicolumn{1}{c}{\textbf{System Prompt}} \\ 
		\midrule
		
		\textbf{Object-Centric} 
		& 
		You are a reasoning-focused vision-language model. You will be given multiple image frames from different viewpoints and a spatial reasoning question. Your goal is to reason step-by-step across these frames before giving the final answer. Please strictly follow the structured reasoning path below. Each stage should be clearly written and numbered. \par\medskip
		
		\textbf{\#\#\# Reasoning Path Template \#\#\#} \newline
		(1) General visual description in each frame \newline
		- Describe what you see in every frame separately. Focus on horizontial spatial relationships, avoid style or non-spatial details. Use consistent object names on the same object. The direction should only discribed in 8 relationships: front, front-right, right, back-right, back, back-left, left, front-left. \newline
		(2) Intermediate reasoning step for connecting cross-frame relationships \newline
		- Describe frame taken position's horizontial spatial relationships based on object spatial relationship observed in (1). \newline
		(3) Query-specific spatial reasoning step \newline
		- Use the above information to perform spatial reasoning about the specific question by fully comprehend the horizontial spatial relationships of all objects in the entire scene. \newline
		(4) Final answer \newline
		- Only output the final choice label in the format: A.$<$answer$>$ or B.$<$answer$>$ or C.$<$answer$>$ or D.$<$answer$>$ \newline
		- Do NOT include any explanation, text, or additional reasoning in this step. \par\medskip
		
		\textbf{\#\#\# Example Output \#\#\#} \newline
		(1) In frame1: the chair is in front of the table. In frame2: the chair is to the right of the table. In frame3: the cell shelf is in front of the table. \newline
		(2) By analyzing the spatial changes of the objects across frames, I establish their relationships: frame2 is taken from the front-right of frame1; frame3 is taken from the front-right of frame2; overall, frame3 is positioned in front of frame1. \newline
		(3) Based on the spatial information in (1) and (2), I construct a complete spatial understanding. From frame2 perspective, the chair is to the right of the table, and the cell shelf is to the left of the table. \newline
		(4) A.left \\ 
		\midrule
		
		\textbf{View-Centric} 
		& 
		You are a reasoning-focused vision-language model. You will be given multiple image frames from different viewpoints and a spatial reasoning question. The provided images are taken at the same position, frame1 is taken by turning 90 degree counter-clockwise from frame0 view, frame2 is taken by turning 180 degree counter-clockwise from frame0 view, frame3 is taken by turning 90 degree clockwise from frame0 view. Your goal is to reason step-by-step across these frames before giving the final answer. Please strictly follow the structured reasoning path below. Each stage should be clearly written and numbered. \par\medskip
		
		\textbf{\#\#\# Reasoning Path Template \#\#\#} \newline
		(1) General visual description in each frame \newline
		- Describe what you see in every frame separately. Focus on horizontial spatial relationships, avoid style or non-spatial details. Use consistent object names on the same object. The direction should only discribed in 8 relationships: front, front-right, right, back-right, back, back-left, left, front-left. \newline
		(2) Intermediate reasoning step for connecting cross-frame relationships \newline
		- We know that the provided images are taken at the same position, frame1 is taken by turning 90 degree counter-clockwise from frame0 view, frame2 is taken by turning 180 degree counter-clockwise from frame0 view, frame3 is taken by turning 90 degree clockwise from frame0 view. \newline
		(3) Query-specific spatial reasoning step \newline
		- Use the above information to perform spatial reasoning about the specific question by fully comprehend the horizontial spatial relationships of all objects in the entire scene. \newline
		(4) Final answer \newline
		- Only output the final choice label in the format: A.$<$answer$>$ or B.$<$answer$>$ or C.$<$answer$>$ or D.$<$answer$>$ \newline
		- Do NOT include any explanation, text, or additional reasoning in this step. \par\medskip
		
		\textbf{\#\#\# Example Output \#\#\#} \newline
		(1) In frame1: the chair is in front of the table. In frame2: the chair is to the right of the table. In frame3: the cell shelf is in front of the table. \newline
		(2) The provided images are taken at the same position, frame1 is taken by turning 90 degree counter-clockwise from frame0 view, frame2 is taken by turning 180 degree counter-clockwise from frame0 view, frame3 is taken by turning 90 degree clockwise from frame0 view. \newline
		(3) Based on the spatial information in (1) and (2), I construct a complete spatial understanding. From frame2 perspective, the chair is to the right of my position, and the cell shelf is to the left of my position. \newline
		(4) A.left \\ 
		
		\bottomrule
	\end{tabularx}
	\caption{System prompts for different viewpoint spatial pattern settings}
	\label{tab:system_prompts}
\end{table*}

\textbf{Overall evaluation settings.} In our evaluations presented in Section 4, all open-source VLMs are executed on a NVidia H100 GPU, and proprietary models are evaluated via their respective commercial APIs using default parameters. A unified set of query templates is used across all models, as summarized in Table~\ref{tab:system_prompts}.

\noindent\textbf{Removal of the view-view category under the view-centric pattern.}
In the view-centric spatial pattern, each rendered view corresponds to a distinct, non-overlapping region of the scene, resulting in no shared objects or geometric anchors across frames. Consequently, models cannot infer inter-viewpoint relationships solely through visual correspondence. To ensure a well-posed reasoning setup, we embed explicit viewpoint cues (i.e., the spatial transformations between frames) directly in the query. Because these cues provide the ground-truth view–view relations, the view–view category is excluded from the view-centric setting.

\section{Explicit Analysis of VLM Spatial Reasoning}
\label{appendix:E}

\subsection{Settings of LLM-as-a-Judge Analysis}
\label{appendix:E1}

\begin{table*}[t]
	\centering
	\footnotesize
	\renewcommand{\arraystretch}{1.35}
	\setlength{\tabcolsep}{8pt}
	
	\begin{tabular}{>{\centering\arraybackslash}m{3.3cm} m{13.7cm}}
		\hline
		\textbf{Component} & \textbf{Judge Prompt Content} \\
		\hline
		
		\textbf{Overall Role} &
		You are an expert geometric reasoning evaluator. Judge the correctness of a VLM’s multi-view spatial reasoning solely using the provided ground truth metadata. The object name in metadata is not exact match with the reasoning path mentioned object name, need to perform semantic matching. The object now semantically matched in the visible object list should not be invovled in the evaluation process.\\
		\midrule
		
		\textbf{Ground-Truth Metadata} &
		(a) Visible object list per frame. \newline
		(b) Inter-frame viewpoint relations: spatial relations among frames. \newline
		(c) Object-viewpoint relations: triplets of the form \verb|<object, direction, object>| using eight canonical directions (front, front-right, right, back-right, back, back-left, left, front-left). \\
		\midrule
		
		\textbf{VLM Reasoning Path} &
		(1) perceptual description. \newline
		(2) cross-frame alignment. \newline
		(3) query-specific spatial inference. \\
		\midrule
		
		\textbf{Evaluation Rules} &
		For reasoning path phase (1): correctness percentage of all stated object spatial relations. \newline
		For reasoning path phase (2): correctness percentage of all stated frame spatial relations. \newline
		For reasoning path phase (3): correctness percentage of all stated spatial relations. \newline
		Do not judge based on final answer or language fluency. \\
		\midrule
		
		\textbf{Required Output Format} &
		Judge must strictly output in JSON \\
		\hline
	\end{tabular}
	
	\caption{LLM-as-a-Judge instruction template.}
	\label{tab:judge_prompt}
\end{table*}

\begin{table*}[t]
	\centering
	\footnotesize
	\renewcommand{\arraystretch}{1.35}
	\setlength{\tabcolsep}{8pt}
	\begin{tabular}{>{\centering\arraybackslash}m{3cm} m{10.5cm}}
		\toprule
		\textbf{Component} & \multicolumn{1}{c}{\textbf{Consistency Evaluation Prompt}} \\
		\midrule
		
		\textbf{Objective} & 
		You will evaluate whether the final answer logically follows from the provided reasoning path. \\
		\midrule
		
		\textbf{Input Data} & 
		(1) The original spatial reasoning query and images. \newline
		(2) The model's multi-phase reasoning steps (Phase 1: frame-wise descriptions; Phase 2: cross-view alignment; Phase 3: query-specific inference). \newline
		(3) The model's final answer choice. \\
		\midrule
		
		\textbf{Output Rules} & 
		Output one of the following labels only: \newline
		\textbf{Consistent}: the final answer directly follows from the reasoning. \newline
		\textbf{Inconsistent}: the reasoning contradicts, fails to support, or is unrelated to the final answer. \\
		\midrule
		
		\textbf{Constraints} & 
		Do NOT revise or regenerate the reasoning. \newline
		Do NOT restate the answer or question. \newline
		Do NOT provide chain-of-thought or explanations. \\
		\bottomrule
	\end{tabular}
	
	\caption{Self-consistency prompting instruction template.}
	\label{tab:consistency_prompt}
\end{table*}

To systematically assess the correctness and stability of model-generated reasoning paths, we adopt an LLM-as-a-judge evaluation framework that is fully aligned with our cognitively grounded four-phase reasoning protocol. For each multi-view VQA instance, we first extract the ground-truth geometric metadata from Blender, including (i) visible object lists for each view (including possible perspective changed view), (ii) all pairwise spatial relations between viewpoints, and (iii) object–view  spatial relations represented as structured view graphs (including possible perspective changed view graph). These graphs provide an explicit and unambiguous geometric reference. The tested VLM is prompted to generate a step-wise reasoning path following our four reasoning phases—perceptual object spatial relationship in frame, frame-level spatial alignment, query-specific mental inference, and final decision. The judge LLM receives three inputs: the VLM’s reasoning text, the ground-truth metadata, and an evaluation instruction specifying how to validate each reasoning step for relational consistency as shown in Table~\ref{tab:judge_prompt}. 

The judge independently scores each three reasoning phase by matching referenced spatial relations against the metadata, excluding the hallucinated entities, and identifying incorrect spatial relationship. We further average scores across multiple judge models including GPT-4o, Claude-4-Sonnet, and Gemini-2.5 Pro to reduce individual model biases. This setup enables phase-wise diagnostic evaluation with early phases primarily test perceptual spatial grounding, while later phases probe cross-view alignment and inferential stability under possible perspective changing. By quantifying degradation patterns across phases, the LLM-as-a-judge framework reveals where spatial reasoning succeeds, where it collapses, and how these behaviors correlate with final answer correctness.

\subsection{VLM Reasoning Paths}
\label{appendix:E2}
\begin{figure}[ht] \centering \includegraphics[width=\linewidth]{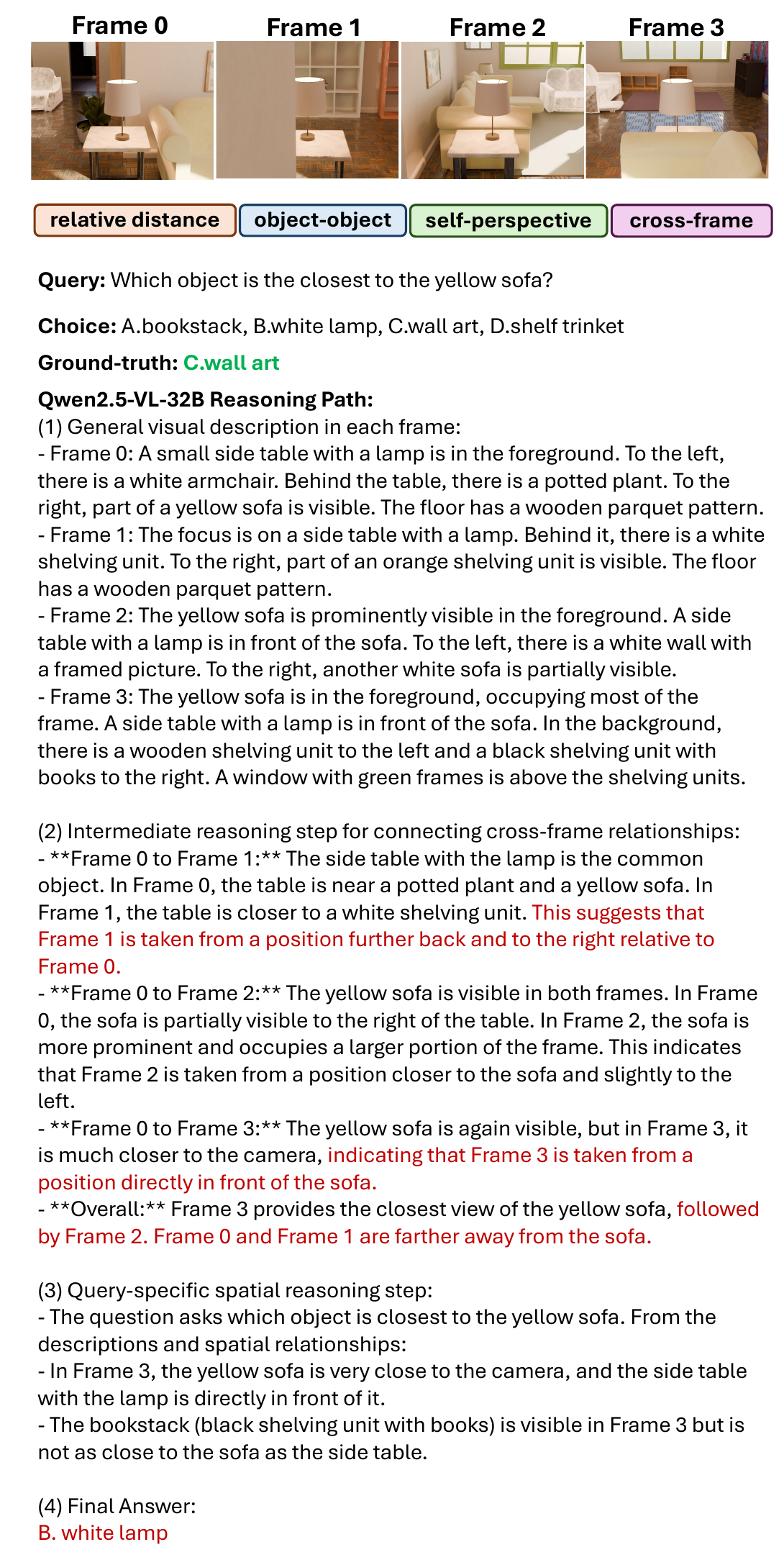} \vspace{-0.25in}\caption{Reasoning path example.} \label{fig:reasoning_path_1} \vspace{-0.1in} \end{figure}

\begin{figure}[ht] \centering \includegraphics[width=\linewidth]{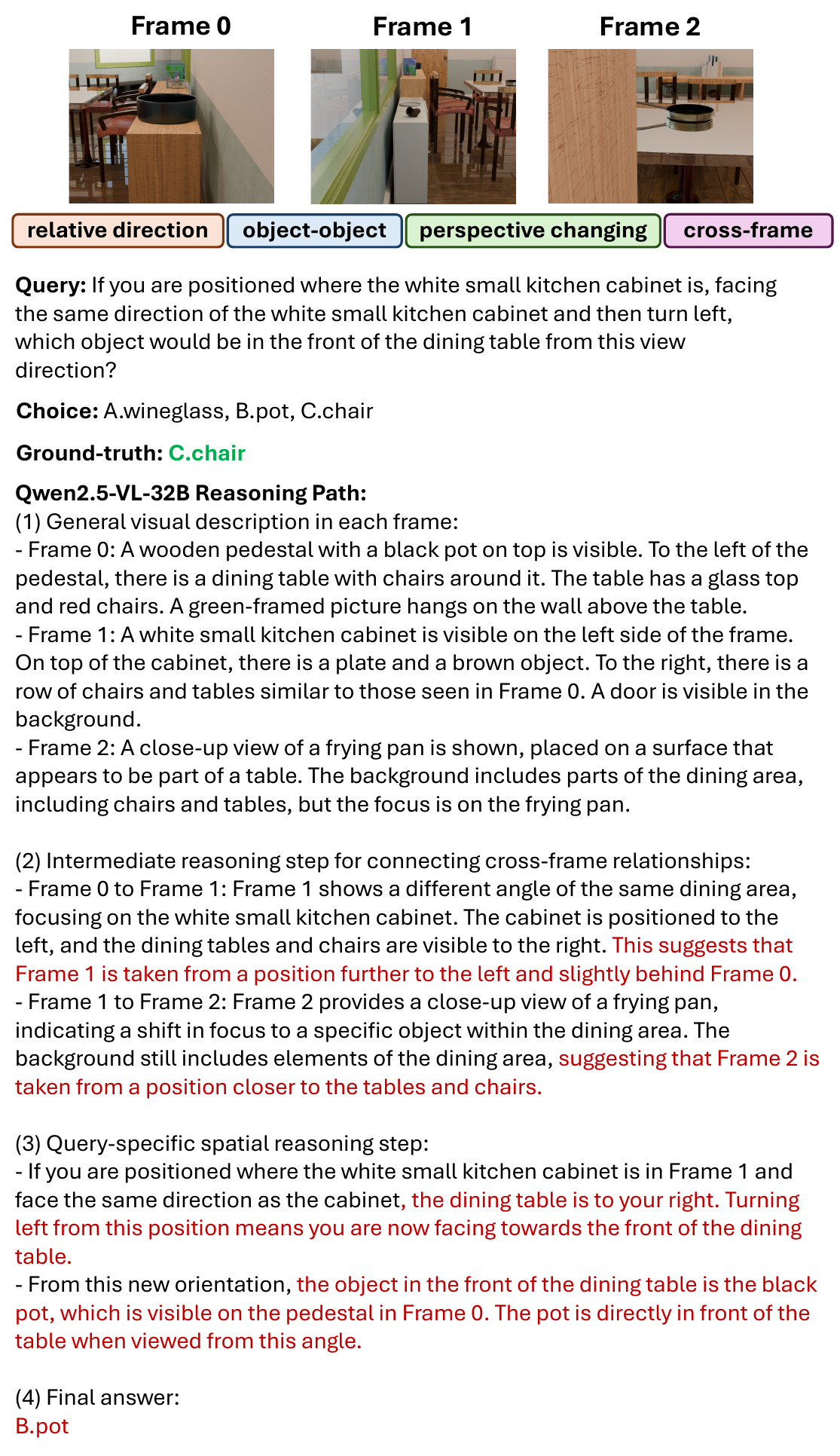} \vspace{-0.25in}\caption{Reasoning path example.} \label{fig:reasoning_path_2} \vspace{-0.1in} \end{figure}

\begin{figure}[ht] \centering \includegraphics[width=\linewidth]{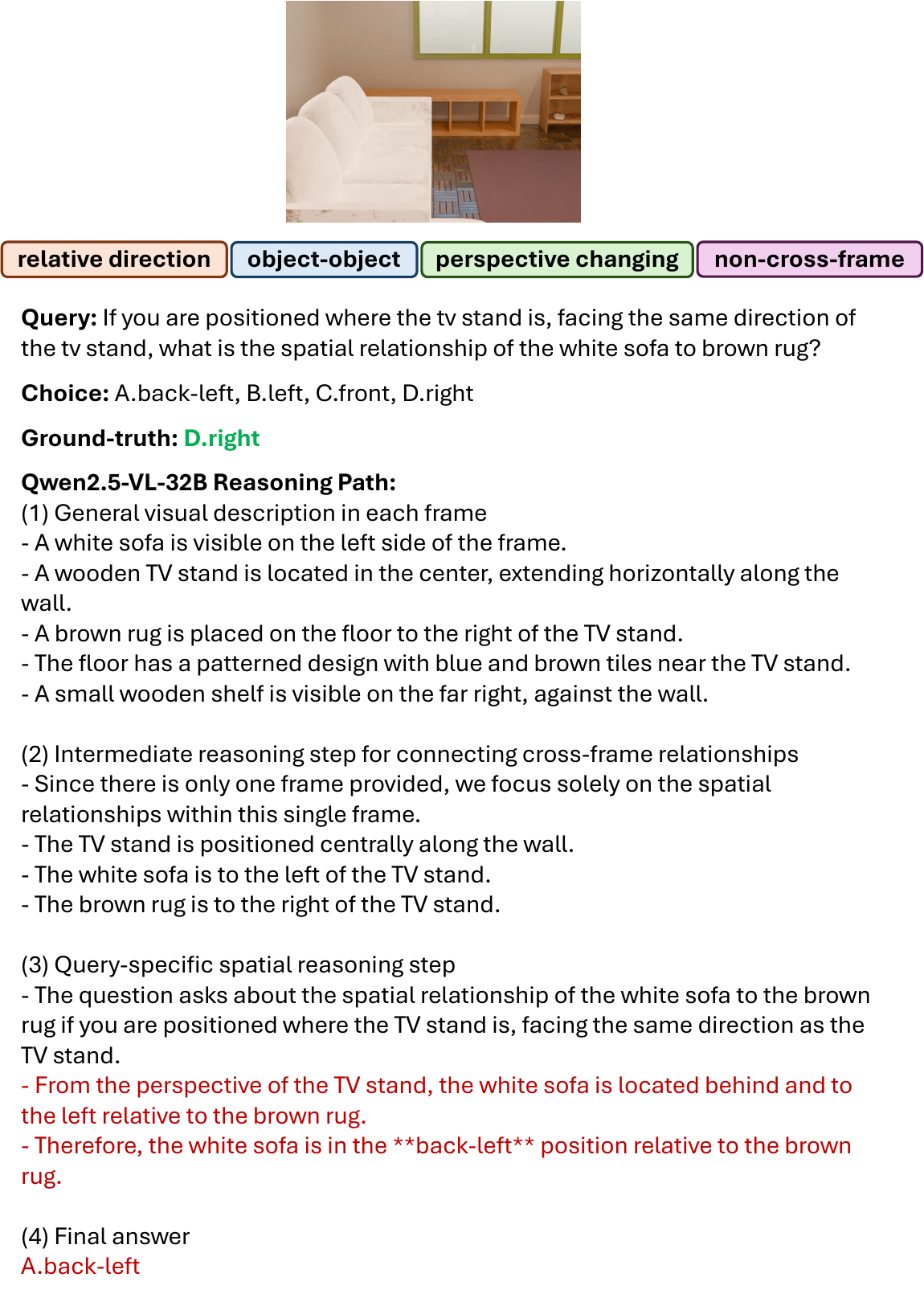} \vspace{-0.25in}\caption{Reasoning path example.} \label{fig:reasoning_path_3} \vspace{-0.1in} \end{figure}

\begin{figure}[ht] \centering \includegraphics[width=\linewidth]{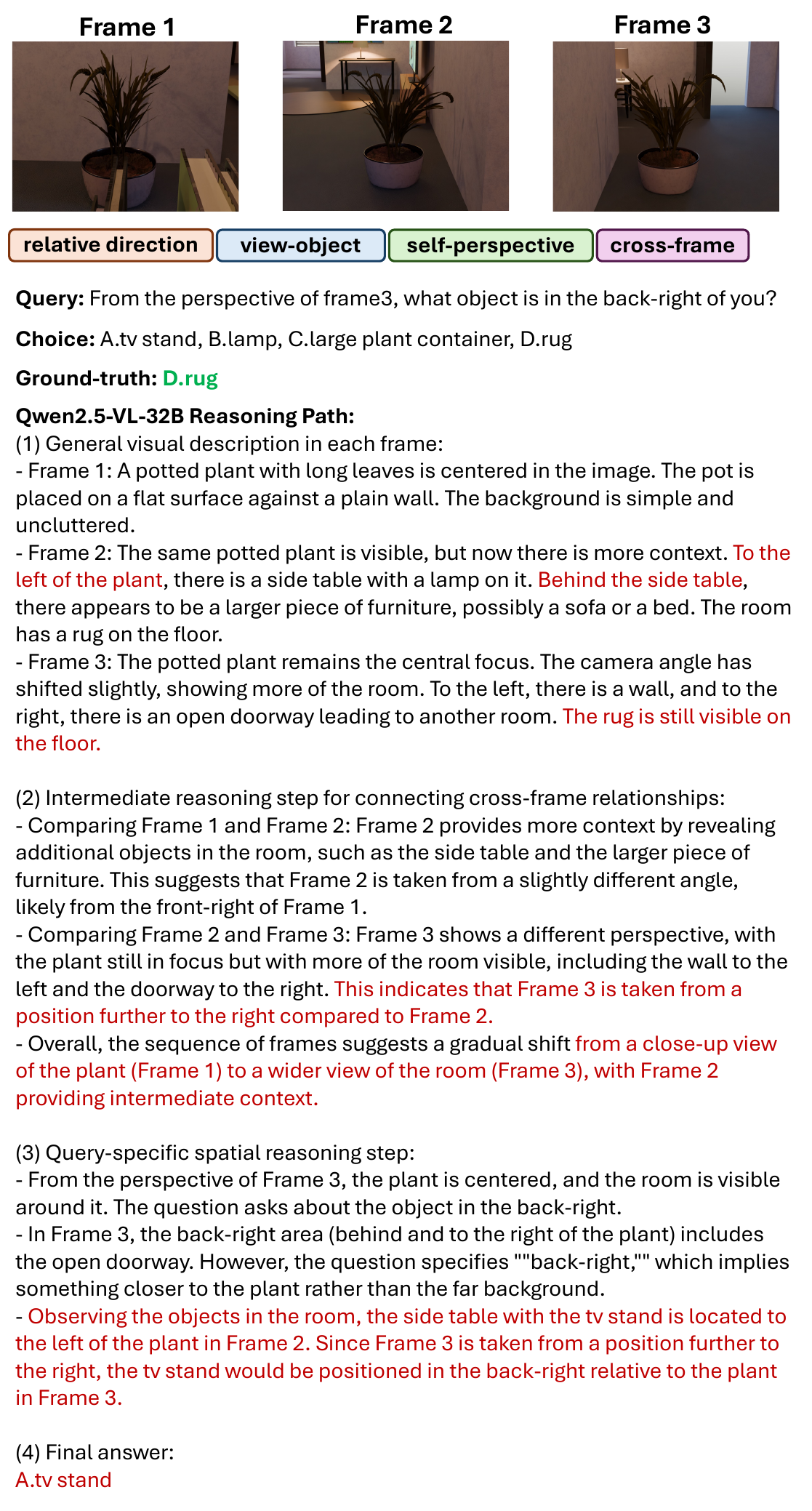} \vspace{-0.25in}\caption{Reasoning path example.} \label{fig:reasoning_path_4} \vspace{-0.1in} \end{figure}

\begin{figure}[ht] \centering \includegraphics[width=\linewidth]{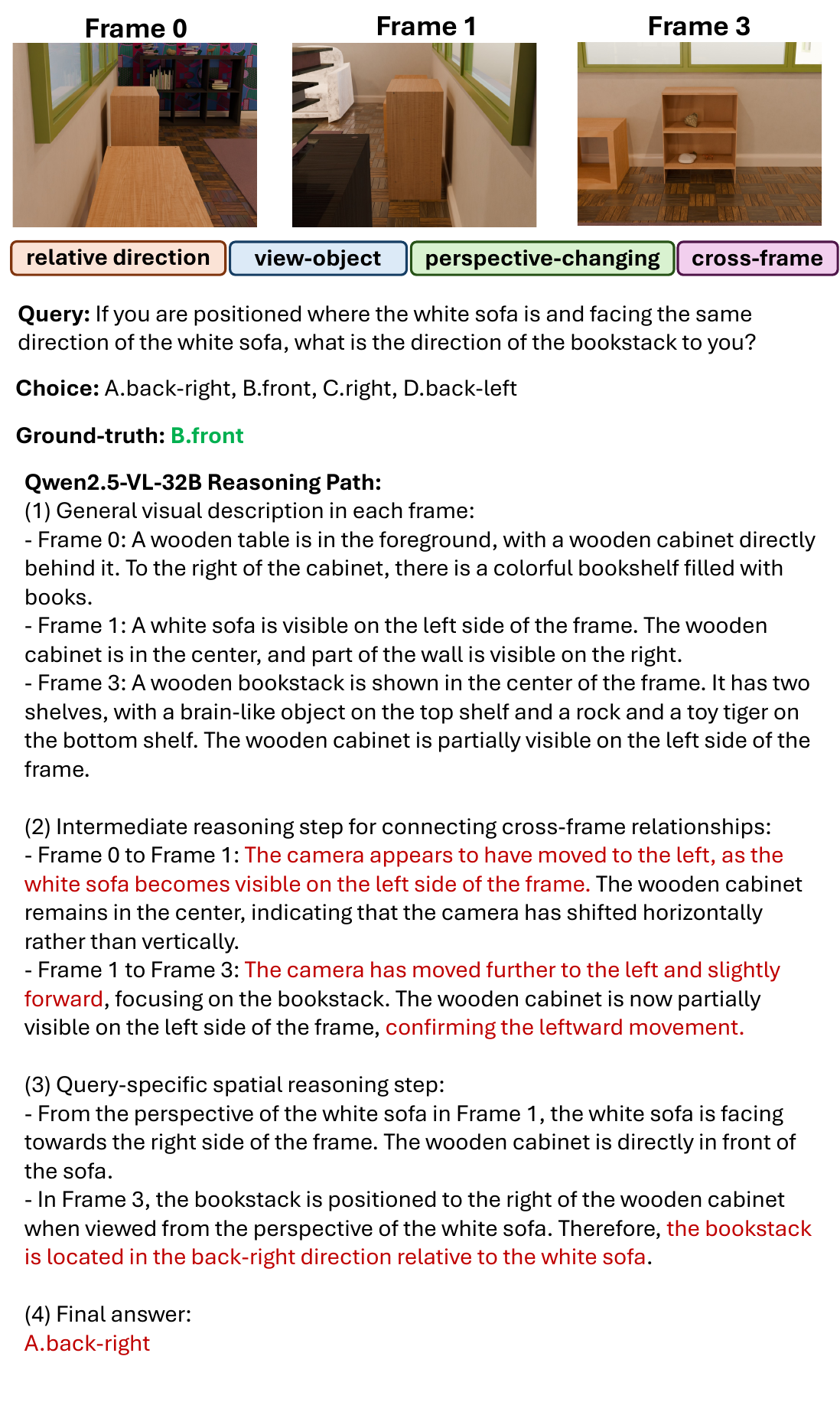} \vspace{-0.25in}\caption{Reasoning path example.} \label{fig:reasoning_path_5} \vspace{-0.1in} \end{figure}

\begin{figure}[ht] \centering \includegraphics[width=\linewidth]{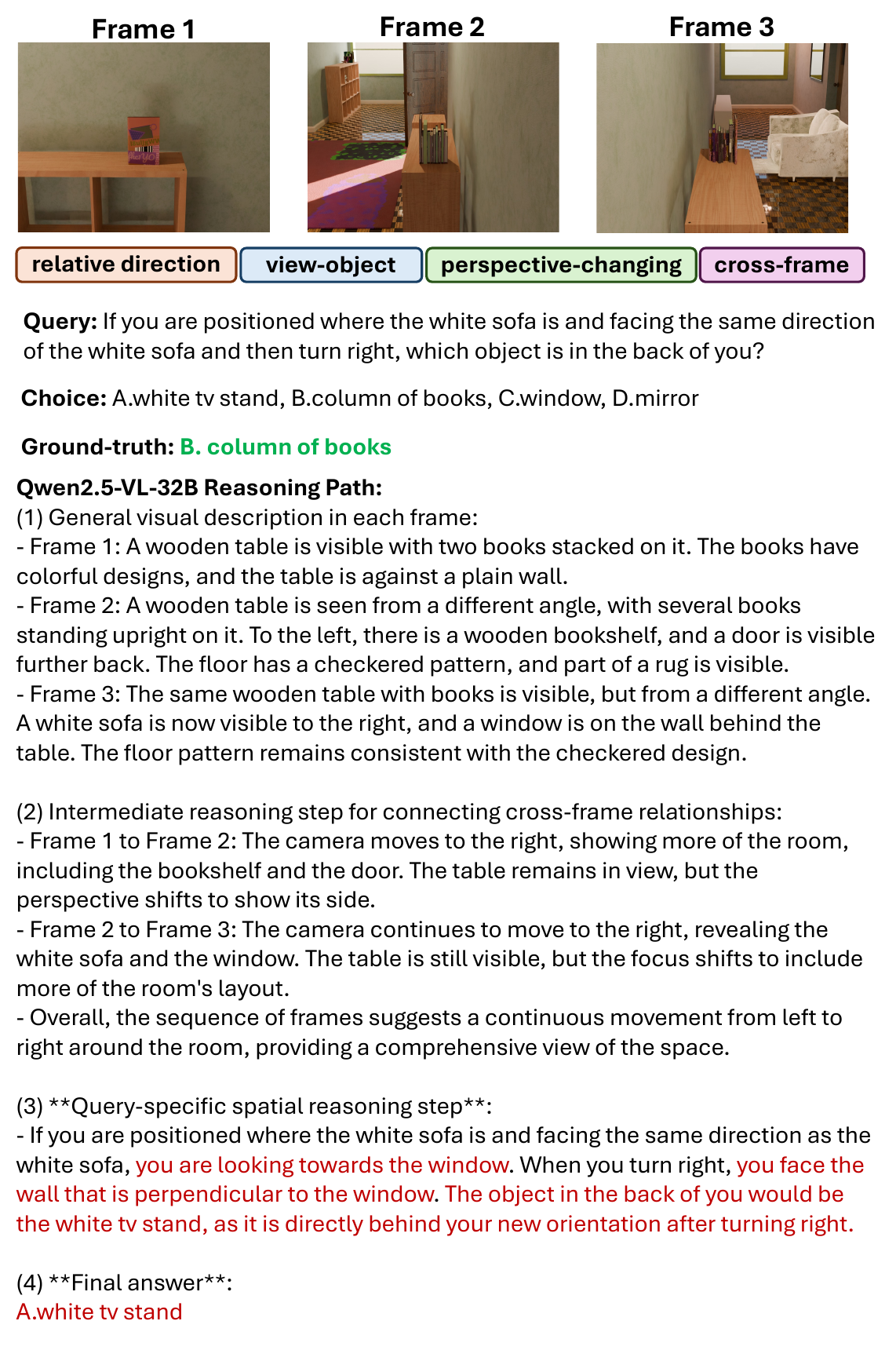} \vspace{-0.25in}\caption{Reasoning path example.} \label{fig:reasoning_path_6} \vspace{-0.1in} \end{figure}

To exemplify the VLM reasoning paths in spatial reasoning failures, we present representative failure cases of the Qwen2.5-VL-32B model across different spatial VQA categories. Figures~\ref{fig:reasoning_path_1}–\ref{fig:reasoning_path_6} illustrate details of incorrectness in reasoning paths (in red) that lead to failure predictions.

\subsection{Details of Self-Consistency Prompting}
\label{appendix:E3}

To assess whether a model’s final prediction is justified by its own multi-phase reasoning trajectory, we employ a self-consistency prompting procedure. Using the instruction template shown in Table~\ref{tab:consistency_prompt} with the spatial VQA query-image set and model multi-phase reasoning path, the same VLM is prompted to judge whether its final answer logically follows from the reasoning it previously generated. The model then returns a binary label, consistent or inconsistent. This evaluation captures the internal coherence between a model’s reasoning path and its final decision, offering a complementary diagnostic signal beyond answer-only accuracy.


\end{document}